\documentclass[runningheads]{llncs}
\usepackage{graphicx}
\usepackage{tikz}
\usetikzlibrary{positioning, calc}
\usepackage{hyperref}
\usepackage{url}
\usepackage{listings}
\usepackage{xspace}
\usepackage{array}
\usepackage{makecell}
\usepackage{graphicx}
\usepackage{multirow}
\usepackage{tcolorbox, soul} 
\usepackage{makecell, multicol, color}
\usepackage{booktabs}
\usepackage{subcaption}
\usepackage{wrapfig}
\usepackage{enumitem}
\usepackage{lipsum}
\usepackage{amsmath,comment}
\usepackage{algorithm}
\usepackage{algpseudocode}

\usepackage{amssymb}
\algtext*{EndWhile} 
\algtext*{EndIf} 
\algtext*{EndFor} 

\usepackage[style=numeric,natbib=true]{biblatex}
\bibliography{reference}
\newtheorem{assumption}[theorem]{Assumption}

\definecolor{codegreen}{rgb}{0,0.6,0}
\definecolor{codegray}{rgb}{0.5,0.5,0.5}
\definecolor{codepurple}{rgb}{0.58,0,0.82}
\definecolor{backcolour}{rgb}{0.95,0.95,0.92}
\lstdefinestyle{mystyle}{
    backgroundcolor=\color{backcolour},   
    commentstyle=\color{codegreen},
    keywordstyle=\color{magenta},
    numberstyle=\tiny\color{codegray},
    stringstyle=\color{codepurple},
    basicstyle=\ttfamily\footnotesize,
    breakatwhitespace=false,         
    breaklines=true,                 
    captionpos=true,                    
    keepspaces=true,                 
    numbersep=5pt,                  
    showspaces=false,                
    showstringspaces=false,
    showtabs=false,                  
    tabsize=2
}

\newcommand{\sys}[0]{Clover\xspace}
\newcommand{\dataset}[0]{CloverBench\xspace}
\newcommand{\dafny}[0]{Dafny\xspace}
\newcommand{\gpt}[0]{GPT-4\xspace}

\newif\ifcomments
\commentstrue
\ifcomments
    \newcommand{\ying}[1]{{\color{blue}{\bf\sf [Ying: #1]}}}
    \newcommand{\todo}[1]{{\color{red}{\bf\sf [TODO: #1]}}}
    \newcommand{\livia}[1]{{\color{orange}{\bf\sf [Livia: #1]}}}
    \newcommand{\clark}[1]{{\color{codegreen}{\bf\sf [Clark: #1]}}}
    \newcommand{\oded}[1]{{\color{purple}{\bf\sf [Oded: #1]}}}
    \newcommand{\edward}[1]{{\color{blue}{\bf\sf [Edward: #1]}}}
    \newcommand{\ben}[1]{{\color{blue}{\bf\sf [Benjamin: #1]}}}
\else
    \newcommand{\ying}[1]{}
    \newcommand{\todo}[1]{}
    \newcommand{\livia}[1]{}
    \newcommand{\clark}[1]{}
    \newcommand{\oded}[1]{}
    \newcommand{\edward}[1]{}
    \newcommand{\ben}[1]{}
\fi

\lstdefinelanguage{Dafny}{
  morekeywords={var, method, ensures, requires, while, if, then,  else, return, returns, decreases, assert, assume, true, false, forall, exists, modifies, invariant, this, new, predicate, reads, lemma},
  morecomment=[l]{//},
  morecomment=[s]{/*}{*/},
  morestring=[b]",
}
\lstdefinestyle{terminal}{
    backgroundcolor=\color{black},
    basicstyle=\footnotesize\color{white}\ttfamily,
    breaklines=true,   
}

\lstdefinestyle{dafnystyle}{
  language=Dafny,
  basicstyle=\ttfamily\small,
  keywordstyle=\color{blue},
  commentstyle=\color{gray},
  stringstyle=\color{red},
  breaklines=true,
  breakatwhitespace=false,
  showstringspaces=false,
}

\title{%
 \vspace*{-1.30cm}
  \sys: \underline{Clo}sed-Loop \underline{Ver}ifiable Code\\ Generation
 \vspace*{-00cm}
}
\toctitle{%
  \sys: \underline{Clo}sed-Loop \underline{Ver}ifiable Code\\ Generation
}

\author{
Chuyue Sun\inst{1}\thanks{Equal Contribution}\orcidID{0009-0005-9226-3688} \quad Ying Sheng\inst{1*}\orcidID{0000-0002-1883-2126}\\ Oded Padon$^2$\orcidID{0009-0006-4209-1635} \quad Clark Barrett$^1$\orcidID{0000-0002-9522-3084}\\
$^1$Stanford University \quad $^2$VMware Research\\
\texttt{\{chuyues, ying1123\}@stanford.edu,
oded.padon@gmail.com,}\\
\texttt{barrett@cs.stanford.edu}
}

\institute{%
  Stanford University \and
  VMware Research
}

\institute{%
}
\authorrunning{Chuyue Sun, Ying Sheng, Oded Padon, Clark Barrett
}

\begin{document}

\setlist[itemize]{noitemsep, topsep=0pt}

\maketitle
\begin{abstract}

The use of large language models for code generation is a
rapidly growing trend in software development. 
However, without effective methods for ensuring the correctness of generated code, this trend could lead to undesirable outcomes. In this paper, we introduce a new approach for addressing this challenge: the \sys paradigm, short for \underline{Clo}sed-Loop \underline{Ver}ifiable Code Generation, which uses consistency checking to provide a strong filter for incorrect code. 
\sys performs consistency checks among code, docstrings, and formal annotations. The checker is implemented using a novel integration of formal verification tools and large language models.
We provide a theoretical analysis to support our thesis that Clover should be effective at consistency checking.
We also empirically investigate its performance on a hand-designed dataset (\dataset) featuring annotated \dafny programs at a textbook level of difficulty. Experimental results show that for this dataset: (i) LLMs are reasonably successful at automatically generating formal specifications; and (ii) our consistency checker achieves a promising acceptance rate (up to $87\%$) for correct instances while maintaining zero tolerance for adversarial incorrect ones (no false positives).
\sys also discovered 6 incorrect programs in the existing human-written dataset MBPP-DFY-50.
\end{abstract}

\section{Introduction}
\label{sec:intro}
Large language models (LLMs) have recently demonstrated remarkable capabilities.  They can engage in conversation, retrieve and summarize vast amounts of information, generate and explain text and code, and much more~\cite{bubeck2023sparks,DBLP:journals/corr/abs-2204-02311,DBLP:journals/corr/abs-2303-08774}. 
Among many possible applications, their ability to synthesize code based on natural language descriptions~\cite{DBLP:journals/corr/abs-2107-03374,DBLP:conf/iclr/ChengX0LNHXROZS23,li2022competition}
is stunning and could potentially enhance the productivity of programmers significantly~\cite{TN22}.  Indeed, futurists are already claiming that in the future, most code will be generated by LLMs (or their successors) and not by humans.

However, there is a fundamental challenge that must be overcome before realizing this future. Currently, there is no trustworthy way to ensure the correctness of AI-generated code \cite{DBLP:journals/corr/abs-2305-01210}. Without some quality control, the prospect of dramatically scaling up code generation is highly concerning and could lead to catastrophic outcomes resulting from faulty code~\cite{DBLP:journals/corr/abs-2308-04451,DBLP:journals/corr/abs-2211-03622,DBLP:conf/uss/SandovalPNKGD23}.
For the most part, the current best practice for curating AI-generated artifacts is to have a human expert in the loop, e.g., \cite{copilot}.  While this is better than nothing, requiring human oversight of AI-generated code limits scalability.  Furthermore,
recent work \cite{hendler2023understanding,DBLP:conf/sp/PearceA0DK22,DBLP:conf/chi/Vaithilingam0G22,DBLP:journals/tosem/XuVN22} confirms the many risks and limitations of using AI even as a code assistant.
Results suggest that developers with access to AI assistants write more insecure code, while at the same time having higher confidence in their code~\cite{DBLP:journals/corr/abs-2211-03622}.

It is becoming clear that curating the quality of AI-generated content will be one of the most crucial research challenges in the coming years. However, in the specific case of generated code, \emph{formal verification} can provide mathematically rigorous guarantees on the quality and correctness of code.  What if there were a way to \emph{automatically} apply formal verification to generated code?  This would not only provide a scalable solution, but it could actually lead to a future in which generated code is \emph{more reliable} than human-written code.

Currently, formal verification is only possible with the aid of time-consuming human expertise.  The main hypothesis of this paper is that \emph{LLMs are well-positioned to generate the collateral needed to help formal verification succeed}; furthermore, they can do this \emph{without compromising the formal guarantees provided by formal methods}.  To understand how, consider the following breakdown of formal verification into three parts: (i) construct a mathematical model of the system to be verified; (ii) provide a formal specification of what the system should do; and (iii) prove that the model satisfies the specification.
For code, step (i) is simply a matter of converting the code into mathematical logic, which can be done automatically based on the semantics of the programming language.  And step (iii) can often be done automatically thanks to powerful automated reasoning systems for Boolean satisfiability (SAT) and satisfiability modulo theories (SMT)~\cite{sathandbook}.  In fact, a number of tools already exist that take a specification (the result of step~(ii)) and some code as input and largely automate steps~(i) and (iii) (e.g., \cite{spark,lattuada2023verus,leino2010dafny}).\footnote{Such tools have plenty of room for improvement and must be extended to more mainstream languages, but separate research efforts are addressing this.}
%
However, step~(ii) appears to be a showstopper for automated formal verification of generated code, as traditionally, significant human expertise is required to create formal specifications and ensure that they are both internally consistent and accurately capture the intended functionality.

Two key insights suggest a way forward. The first insight is simply a shift in perspective: the result of any AI-based code generation technique should aim to include \emph{not only code, but also formal specifications}.  The second insight is that given these components (and a description in natural language), we can use formal tools coupled with generative AI techniques to \emph{check their consistency}.
We name our approach \emph{Clover}, short for \emph{Closed-loop Verifiable Code Generation}, and we predict that \sys, coupled with steadily improving generative AI and formal tools, will enable a future in which fully automatic, scalable generation of formally verified code is feasible. This paper charts the first steps toward realizing this vision.

\begin{wrapfigure}{tr}{0.4\textwidth}
\vspace{-1.5em}
\includegraphics[width=0.4\textwidth]{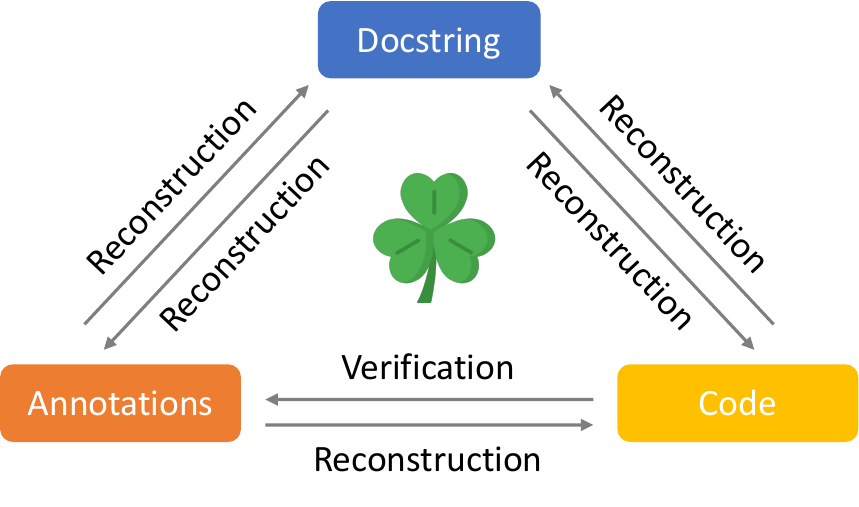}
\vspace{-2em}
\footnotesize \caption{The \sys paradigm}
\label{fig:overview}
\vspace{-0.5em}
\end{wrapfigure}

The \sys paradigm consists of two phases: generation and verification.
In this paper, we also assume that a precise natural language description of the desired functionality is available.
In the first (generation) phase, some process is used to create code annotated with formal specifications.
For simplicity, we refer to the formal specifications as ``annotations'' and the natural language descriptions as ``docstrings'' going forward.  
It is worth noting that, in other scenarios, including annotating an existing codebase or generating code given specifications, one or two of these components (code, annotations, docstrings) might already exist, in which case generative AI might be used to construct only the other(s).
In fact, the second phase is completely agnostic to the process used in the first phase; we simply insist that the result of the first phase has all three components: code, annotations, and docstrings.
In the second (verification) phase, a series of \emph{consistency checks} are applied to the code, annotations, and docstrings (see Figure~\ref{fig:overview}).  The Clover hypothesis is that if the consistency checks pass, then (i) the code is functionally correct with respect to its annotations; (ii) the annotations capture the full functionality of the code; and (iii) the code and its annotations also align with natural language descriptions of the functionality (docstrings).

The idea is that we can unleash increasingly powerful and creative generative AI techniques in the generation phase, and then use the verification phase as a strong filter that only approves of code that is formally verified, accurately documented, and internally consistent.

In this paper, we focus on the verification phase, though we also include some demonstrations of the generation phase in our evaluation.  Our contributions include:%
\footnote{In addition, a theoretical framework which argues for the trustworthiness of the Clover approach is available in \cite[Appendix A.1]{sun2024clover}. }
\begin{itemize}[topsep=0pt, leftmargin=16pt]
    \item the Clover paradigm with a solution for the verification phase (Section~\ref{sec:clover_schema});
    \item the \dataset dataset, featuring manually annotated Dafny programs with docstrings, which contains both ground-truth examples and adversarial incorrect examples (Section~\ref{sec:dataset});
    \item a demonstration of the feasibility of using \gpt to generate code, specifications, and both (Section~\ref{sec:exp:generation});
    \item implementation and evaluation of the verification phase of the \sys paradigm using \gpt and the Dafny verification tool (Section~\ref{sec:exp:ground_truth}, \ref{sec:exp:mbpp-dafny}, and~\ref{sec:exp:inconsistency}).
\end{itemize}

Our initial results on \dataset are promising.  Our implementation accepts $87\%$ of the correct examples and rejects $100\%$ of the adversarial incorrect examples.
We expect that the acceptance rate can be improved in a variety of ways while maintaining the strong ability to reject incorrect code. Beyond \dataset, \sys also correctly detects 6 incorrect programs and accepts $89\%$ of the correct programs in the external dataset MBPP-DFY-50~\cite{misu2024towards}.
\footnote{The \dataset dataset and \sys consistency checking implementation will be made available after the anonymous review period.}

\section{Preliminaries: Deductive Program Verification}
\label{sec:prelim}



Deductive program verification provides a framework for mathematically proving that programs are correct~\cite{Floyd67,Hoare69}.
A standard approach is to first \emph{annotate} code with preconditions, postconditions, and loop invariants, and then check that the code satisfies the specification given by these annotations. That is, if the code is executed starting from a program state that satisfies the precondition, the resulting program state after executing the code will satisfy the postcondition. Checking whether a given piece of code meets the specification corresponding to some set of annotations can be done by checking the validity of logical formulas known as verification conditions, which is typically done automatically using satisfiability modulo theories (SMT) solvers. \dafny is a programming language used in our evaluation with state-of-the-art support for deductive verification~\cite{leino2010dafny}.  \dafny's back-end includes both a compiler, capable of generating a runnable binary, and a verifier, which formally checks whether the code conforms to its specification.  


In this paper, we assume annotations are given at the function level. For example, a function for finding the maximal element in an array of integers will have a precondition requiring that the input array is nonempty, and a postcondition ensuring that the return value is indeed the maximal element of the input array. Loops must be accompanied by loop invariants,
which are used for a proof by induction on the number of loop iterations. 
For example, Listing~\ref{lst:maxarray} shows a Dafny function for finding the maximal element of an array, with a docstring, a precondition, two postconditions, and a loop invariant. Dafny is able to automatically verify this function with respect to these formal annotations.

\begin{lstlisting}[style=dafnystyle,basicstyle=\ttfamily\scriptsize,label=lst:maxarray,caption={Dafny function with consistent code, docstring, and annotations.}]
// Find the maxiaml element in an integer array
method maxArray(a: array<int>)returns (m: int)
 requires a.Length>=1
 ensures exists k :: 0<=k<a.Length && m==a[k]
 ensures forall k :: 0<=k<a.Length ==> m>=a[k] 
{
 m := a[0];
 var i := 1;
 while (i < a.Length)
 invariant 0<=i<=a.Length &&
           (forall k :: 0<=k<i ==> m>=a[k]) &&
           (exists k::0<=k<i && m==a[k])
 {
  m := if m>a[i] then  m else a[i];
  i := i + 1;
 }
}
\end{lstlisting}
\hfill

\begin{lstlisting}[basicstyle=\ttfamily\scriptsize,label=lst:gendoc,caption={Example of generated docstring.}]
"This method returns the maximum value, m, in the integer array a, ensuring that m is greater than or equal to all elements in a and that m is indeed an element of a"
\end{lstlisting}
\vspace{-1em}
\begin{lstlisting}[style=dafnystyle,basicstyle=\ttfamily\scriptsize,label=lst:genanno,caption={Example of generated annotations.}]
requires a.Length > 0;
ensures forall k::0<=k<a.Length ==> a[k]<=m
ensures exists k::0<=k<a.Length && a[k]==m
\end{lstlisting}
\vspace{-1em}
\begin{lstlisting}[style=dafnystyle,basicstyle=\ttfamily\scriptsize,label=lst:gencode,caption={Example of generated code (loop invariant omitted).}]
 var i := 0;
 m := a[0];
 while i<a.Length { 
  if (a[i] > m) { m := a[i]; }
  i := i+1;
 }
\end{lstlisting}

\section{\sys}
\label{sec:clover}
\subsection{\sys Generation Phase}
\label{sec:clovergen}

As mentioned in Section~\ref{sec:intro}, \sys expects the output of the generation phase to consist of code, annotations, and docstrings.  These could be generated in a variety of ways.  In this paper, we include a feasibility study for three possible instances of the generation phase.  

First, we consider the case where the annotations (i.e. the formal specifications) are provided, and an LLM is asked to generate the code.  This is analogous to the standard synthesis problem that is well-studied in PL research~\cite{DBLP:journals/ftpl/GulwaniPS17}\cite{manna1975knowledge}\cite{manna1971toward}\cite{waldinger1969prow}.

Second, we explore the opposite: generating annotations given the code.  This use case could be relevant for someone trying to verify legacy code.

Finally, we explore the possibility of generating both code and annotations from a precise natural language description.  This use case aligns with our proposed vision that LLMs should include specifications when generating code from natural language.

Our goal with these evaluations is not necessarily to chart new research directions, as all of these directions are worthy of a much more targeted research effort (and indeed, there are many such efforts underway~\cite{brandfonbrener2024verified,misu2024towards,10.1117/12.3011627}).  Rather, our goal here is simply to demonstrate the feasibility of different instances of the generation phase in order to lend credibility to the overall \sys vision. 
We report on an evaluation of each of these use cases in Section~\ref{sec:evaluation}.

\subsection{\sys Verification Phase}
\label{sec:clover:consistency-check}

As mentioned in Section~\ref{sec:intro}, \sys expects the input of the verification phase to contain three components: code, annotations, and docstrings. \emph{Additionally, we expect that each of the three components provides sufficient detail to unambiguously determine a unique result of running the code on any given input.}
The verification phase checks the consistency of every pair of components, as shown in Figure~\ref{fig:overview}, and succeeds only if all checks pass.
Docstrings and annotations are consistent if they contain the same information, i.e., they imply each other semantically.
The notion of consistency between a docstring and code is similar.
On the other hand, to assess the consistency between code and annotations, we can leverage deductive verification tools.


\begin{figure}
\begin{algorithm}[H]
        \caption{\sys Consistency Check ($k=1$) \label{algo:consistency-check}}
        \begin{algorithmic}
\Require Docstring $d$, annotations $a$, code $c$.
\Ensure $True/False$

\State

\State Set number of tries $m=3$ 
\If{Dafny fails to verify $a, c$} \Comment{annotation soundness}
    \State \textbf{Return} $False$
\EndIf
\For{$i$ = 1 to $m$} \Comment{annotation completeness}
    \State Call LLM to generate code $c'$ from $a$.
    \If{$c'$ successfully compiles}
        \State{break}
    \Else
        \State{Provide feedback from failed compilation to LLM}
    \EndIf
\EndFor
\If{$c'$ is not equivalent to $c$}
    \State \textbf{Return} $False$
\EndIf
\For{$i$ = 1 to $m$}\Comment{doc2code}
    \State Call GPT-4 to generate code $c'$ from $d$.
    \If{$c'$ successfully compiles}
        \State{break}
    \Else
        \State{Provide feedback from failed compilation to LLM}
    \EndIf
\EndFor
\If{$c'$ is not equivalent to $c$}
    \State \textbf{Return} $False$
\EndIf
\For{$i$ = 1 to $m$}\Comment{code2doc}
    \State Call GPT-4 to generate docstring $d'_i$ from $c$.
\EndFor
\If{all $d'_i$ are not equivalent to $d$}
    \State \textbf{Return} $False$
\EndIf

\For{$i$ = 1 to $m$}\Comment{doc2anno}
    \State Call GPT-4 to generate annotations $a'$ from $d$.
    \If{$a'$ successfully compiles}
        \State{break}
    \Else
        \State{Provide feedback from failed compilation to LLM}
    \EndIf
\EndFor
    \If{$a'$ is not equivalent to $a$}
        \State \textbf{Return} $False$
    \EndIf
\For{$i$ = 1 to $m$}\Comment{anno2doc}
    \State Call GPT-4 to generate docstring $d'$ from $a$.
\EndFor
\If{all $d'$ are not equivalent to $d$}
    \State \textbf{Return} $False$
\EndIf        
\State \textbf{Return} $True$
\end{algorithmic}
    \end{algorithm}
\end{figure}

\label{sec:clover_schema}

One key idea used to check consistency between components in Figure~\ref{fig:overview} is \emph{reconstruction testing}. Given the three components (code, docstring, annotations) as input, 
we try to reconstruct a single component from a single other component, and then we check if the reconstructed result is equivalent to the original component. We do this for five out of the six (directed) edges of Figure~\ref{fig:overview}. A special case is checking that the code conforms to the annotations, where we use formal verification based on deductive verification tools instead of a reconstruction test.
For an input instance to pass the verification phase, it must pass all six tests.
For the reconstruction itself, we use an LLM (our evaluation uses GPT-4), and for equivalence testing, we use LLMs to compare text, formal tools to compare annotations, and pointwise sampling to compare code.
A running example is provided in Section~\ref{running-example}.
Listings~\ref{lst:gendoc},~\ref{lst:genanno}, and~\ref{lst:gencode} are examples of generated artifacts.
We explain how these checks are done in detail next.  Pseudocode is shown in Algorithm~\ref{algo:consistency-check}.
\footnote{For more discussion about limitations and variants of, and future directions for \sys, see \cite[Appendix A.4]{sun2024clover}.}


\textbf{Code-Annotations Consistency}
%
(1. Code $\rightarrow$ Annotations: Soundness) A deductive verification tool (our evaluation uses Dafny) checks that the code satisfies the annotations. This is a standard formal verification check (see Section~\ref{sec:prelim}).
(2. Annotations $\rightarrow$ Code: Completeness)
To prevent annotations that are too trivial from being accepted, we test whether the annotations are strong enough by testing if they contain enough information to reconstruct functionally equivalent code.  Given the annotations, we use an LLM to generate new code.
Then, we check the equivalence between the generated and the original code.
If the equivalence check passes, the annotations are considered complete.

\textbf{Annotation-Docstring Consistency}
    (1. Annotations $\rightarrow$ Docstring)
    An LLM is asked to generate a new docstring from the annotations. Then, the new and the original docstrings are checked for semantic equivalence.
    (2. Docstring $\rightarrow$ Annotations)
    An LLM is asked to generate new annotations from the docstring. Then, the new and the original annotations are checked for logical equivalence.

\textbf{Code-Docstring Consistency}
    (1. Docstring $\rightarrow$ Code)
    An LLM is asked to generate code from the docstring. Then, the new and the original code are checked for functional equivalence.
    (2. Code $\rightarrow$ Docstring)
    An LLM is asked to generate a new docstring from the code. Then, the new and the original docstrings are checked for semantic equivalence.

We consider the methods used for equivalence checking to be parameters to \sys.  We discuss some possibilities (including those used in our evaluation) below.

\textbf{Equivalence Checking for Code}
Standard equivalence checks for code include input-output comparisons, concolic testing (\cite{DBLP:conf/osdi/CadarDE08,DBLP:journals/cacm/CadarS13,DBLP:journals/cacm/King76,DBLP:conf/pldi/UdupaRDMMA13}), and even full formal equivalence checking (e.g., \cite{DBLP:conf/pldi/ChurchillP0A19}).
Our evaluation checks that the outputs agree on a set of inputs included as part of the \dataset dataset. This test is, of course, imprecise
, but our evaluation suggests that it suffices for the level of complexity in \dataset. 
For example, the generated code of Listing~\ref{lst:gencode} is equivalent to the original code in Listing~\ref{lst:maxarray}, and indeed our equivalence check succeeds for this example.
More advanced equivalence checking techniques might be required for more complex examples.

\textbf{Equivalence Check for Docstrings}
Checking equivalence between docstrings is challenging, as natural language is not mathematically precise.
In our evaluation, we ask an LLM (\gpt) to check whether two docstrings are semantically equivalent.  
For example, it accepts Listing~\ref{lst:gendoc} as equivalent to the docstring in Listing~\ref{lst:maxarray}.
Other NLP-based semantic comparisons may also be worth exploring.


\textbf{Equivalence Check for Annotations} 
To check the equivalence of two sets of annotations, we write the equivalence as a formal lemma and ask a formal tool (in our evaluation, we again use Dafny) to prove the lemma.
This method is sound in the sense that it succeeds only if the two sets of annotations are indeed equivalent. 
For example, we are able to automatically prove that the annotations in Listing~\ref{lst:genanno} are equivalent to those in Listing~\ref{lst:maxarray}.
Note that this process may fail, even on equivalent annotations, due to the limitations of the verification tool being used. 
The specific equivalence checking template we use is described in Section~\ref{sec:dataset} and is included as part of our \dataset dataset.

Although there are many approximate approaches, the two parts that leverage formal tools, the soundness check and the equivalence check for annotations, are exact.
The equivalence check used for code is also strong, though not perfect.
\textit{These checks strongly contribute to the lack of false positives in our evaluation.}%
\footnote{An analytical model of reconstruction tests is provided in \cite[A.1]{sun2024clover}.}

\subsection{Consistency Checking Example}
\label{running-example}
For illustration purposes, before the evaluation section, we describe how each step described above is carried out for the \texttt{maxArray} example (Listing~\ref{lst:maxarray}).
\begin{lstlisting}[style=dafnystyle,basicstyle=\ttfamily\scriptsize,label=lst:annoinput,caption={Annotation Input}]
method foo(a: array<int>) returns (m: int)
 requires a.Length >= 1
 ensures (forall k :: 0<=k<a.Length ==> m>=a[k]) && (exists k :: 0<=k<a.Length && m==a[k])
{
//TOFILL
}
\end{lstlisting}

\begin{minipage}{0.48\textwidth}
\begin{lstlisting}[style=dafnystyle,basicstyle=\ttfamily\scriptsize,label=lst:codeinput,caption={Code Input}]
method foo(a: array<int>) returns (m: int)
//TOFILL
{
 m := a[0];
 var i := 1;
 while (i < a.Length)
 invariant 0<=i<=a.Length &&
           (forall k :: 0<=k<i ==> m>=a[k]) &&
           (exists k :: 0<=k<i && m==a[k])
 {
  m := if m>a[i] then  m else a[i];
  i := i + 1;
 }
}
\end{lstlisting}
\end{minipage}
\hfill
\begin{minipage}{0.48\textwidth}
\begin{lstlisting}[style=dafnystyle,basicstyle=\ttfamily\scriptsize,label=lst:gen_code,caption={Generated code}]
method foo(a: array<int>) returns (m: int)
{
  var i := 0;
  m := a[0];
  while i<a.Length
  {
    if(a[i] > m) {
      m := a[i];
    }
    i := i+1;
  }
}
\end{lstlisting}
\end{minipage}

\begin{lstlisting}[style=dafnystyle,basicstyle=\ttfamily\scriptsize,label=lst:docinput,caption={Docstring Input}]
// specification: Returns the maximum value m present in the array a.
method maxArray(a: array<int>) returns (m: int)
//TOFILL
\end{lstlisting}

\textbf{anno-sound}
Soundness is checked by simply running the \dafny verifier on the annotated code shown in Listing~\ref{lst:maxarray}.

\textbf{anno-complete}
\label{sec:anno_complete}
For the annotations to be complete with respect to the code, we must be able to reconstruct the code from the annotations alone. 
Therefore, we ask \gpt to generate code from the anonymized function signature and the annotations (Listing~\ref{lst:annoinput}).
In the prompt, we instruct the LLM to generate code based on the \dafny specification in natural language, without providing any few-shot examples.
We run and provide feedback from the \dafny compiler up to three times to help \gpt fix its code generation.  For this example, \gpt generates the correct code on the first try, shown in Listing~\ref{lst:gen_code}.
Then, we check if the generated code is equivalent to the original ground-truth code by comparing their outputs.\footnote{Example code for this check is shown in \cite[Appendix A.8]{sun2024clover}.}


\textbf{doc2anno}
We try to reconstruct equivalent ground-truth annotations from the docstring alone. First, we call \gpt with the docstring and the function signature (Listing~\ref{lst:docinput}) asking for the annotations.
To eliminate simple syntax errors, we try to compile the generated annotations with an empty code body and use error messages generated by the \dafny compiler as feedback (up to 3 times). Results presented in Section~\ref{sec:exp:generation}, above, suggest that the feedback mechanism is quite important.  For our example, \gpt generates correct annotations on the first try, shown in Listing~\ref{lst:gen_anno}.
\begin{lstlisting}[style=dafnystyle,basicstyle=\ttfamily\scriptsize,label=lst:gen_anno,caption={Generated annotations}]
  requires a.Length > 0;
  ensures forall k :: 0 <= k < a.Length ==> a[k] <= m;
  ensures exists k :: 0 <= k < a.Length && a[k] == m;
\end{lstlisting}

\textbf{anno2doc}
\label{sec:anno2doc}
To reconstruct a docstring from annotations, we ask \gpt to generate a new docstring three times independently in one session in plain natural language. If one of them is equivalent to the original docstring, the check passes.
We consider two docstrings to be equivalent if they contain the same information about the functional behavior of the program, ignoring implementation details that do not affect functionality. In the prompt, we ask, ``Do these two docstrings describe the exact same functional behavior of a \dafny program? Return 'Yes' or 'No'.'' followed by the two docstrings in question (see GPT-4 System Prompt in \cite[Appendix A.7]{sun2024clover}). Note that the two calls to \gpt are independent to ensure that the second call contains no memory of the first call. That is, the answer to the question of whether the original and the generated docstrings are semantically equivalent is unaffected (other than by bias inherent in the model) by the first call to generate an equivalent docstring from the original.
For our example, \gpt generates a correct docstring on the first try, shown below:
\begin{quote}
\small
This method returns the maximum value, m, in the integer array a, ensuring that m is greater than or equal to all elements in a and that m is indeed an element of a.
\end{quote}
\textbf{code2doc}
The process is almost identical to \verb|anno2doc|. The only difference is that in order to ensure the code provides all the information needed for the docstring generation, we embed the preconditions into the code in the form of \verb|assert| statements.

\textbf{doc2code}
This process leverages one of the most common use cases of \gpt: generating code from a natural language description. The concrete steps are similar to that described in \verb|anno-complete|. The only difference is that instead of using verifier-generated error messages, we use compiler-generated error messages since we want to ensure that the code generation relies only on the docstring.

\section{Evaluation}
\label{sec:evaluation}

We have implemented a first prototype of our \sys consistency checking algorithm using \gpt~\cite{DBLP:journals/corr/abs-2303-08774} as the LLM and using the \dafny programming language and verification tool~\cite{leino2010dafny}.  We selected \dafny because it provides a full-featured and automatic deductive verification toolkit including support for a rich language of formal specifications and a backend compiler linking to a verifier.  But \sys can be instantiated using any language and tool supporting deductive program verification.  Note that it is also crucial that the selected LLM has a good understanding of the programming language.  In our case, we were pleasantly surprised to discover that \gpt understands \dafny programs well enough to perform the translations between code, docstrings, and annotations that \sys relies on (Section~\ref{sec:exp:generation}), despite the fact that \dafny is not a mainstream programming language.
In our evaluation, we use \dafny version $4.0.0.50303$ with Z3 version $4.8.12$.  The evaluation also uses a concrete set of \dafny examples which we describe next.


\subsection{Dataset: \dataset}
\label{sec:dataset}

\subsubsection{\dafny}

There have been several popular datasets for code generation in different domains \cite{DBLP:journals/corr/abs-2108-07732,DBLP:journals/corr/abs-2107-03374,DBLP:conf/nips/HendrycksBKMAGB21,DBLP:conf/acl/YinLXRWSHBCMPS23,DBLP:conf/icml/Lai0WZZZYFWY23}, but none of them contain annotations or use the \dafny language.
Furthermore, we wanted to carefully curate the programs used to test our first \sys prototype.  In particular, as mentioned above, we require the docstring and annotations to precisely specify a unique output for every input.
For these reasons, we introduce a new hand-crafted dataset we call \dataset. We expect to add and improve it over time, but
at the time of writing, it is based on 60 small hand-written example programs as might be found in standard CS textbooks.%
\footnote{Since we wanted to concentrate on the most basic scenario initially, our initial dataset only features examples containing exactly one method and no helper functions.}
For each program, there are five variants: a ``ground-truth'' variant whose code, annotations, and docstring are correct and consistent (verified by hand); and 4 adversarial incorrect variants.
Associated with each example, there is also one set of inputs and one \dafny code template for annotation equivalence checking.
We discuss possible data contamination issues in \cite[Appendix A.4]{sun2024clover}.

It is worth noting that recently, independent and concurrent work \cite{brandfonbrener2024verified,misu2024towards} on \dafny annotation generation has produced some \dafny examples with annotations that are similar to \dataset.  However, there are only a limited number of these benchmarks, and they do not always meet the strict criteria we have imposed in this paper (single-method code with precise specifications), and thus our carefully curated \dataset is still needed.  In MCTS~\cite{brandfonbrener2024verified}, only 5 examples are provided. In dafny-synthesis~\cite{misu2024towards}, the authors translate some programs from MBPP~\cite{DBLP:journals/corr/abs-2108-07732}, a data set of Python programs, into \dafny. We do evaluate \sys on a subset of these benchmarks in Section~\ref{sec:exp:ground_truth}, below.

\textbf{Set of Inputs}
Each program in \dataset contains five individual tests designed to run that program on a specific input value. We use these tests as a rough check for whether a piece of generated code is equivalent to the original code. 
If the generated code has the same output as the original code for all five tests, then the code is considered to be equivalent (See \cite[Appendix A.8]{sun2024clover}).

\textbf{Annotation Equivalence Checking Template}
Each template can be used to formally verify the consistency of two sets of annotations with \dafny. For two sets of annotations 
$a$ and $b$ to be equivalent, the preconditions and postconditions of $a$ and $b$ must be verified to be equivalent separately.  We use a script to automatically create annotation templates.\footnote{Details and an example are shown in \cite[Appendix A.7]{sun2024clover}.}


\subsection{Generation Phase}
\label{sec:exp:generation}
As mentioned in Section~\ref{sec:clovergen}, we explored three use cases for the generation phase.  In all cases, we use \gpt as the generating LLM.

\begin{figure}[ht]
\vspace{-0.5em}
\begin{subfigure}[b]{0.3\textwidth}
    \centering
    \includegraphics[width=1\textwidth]{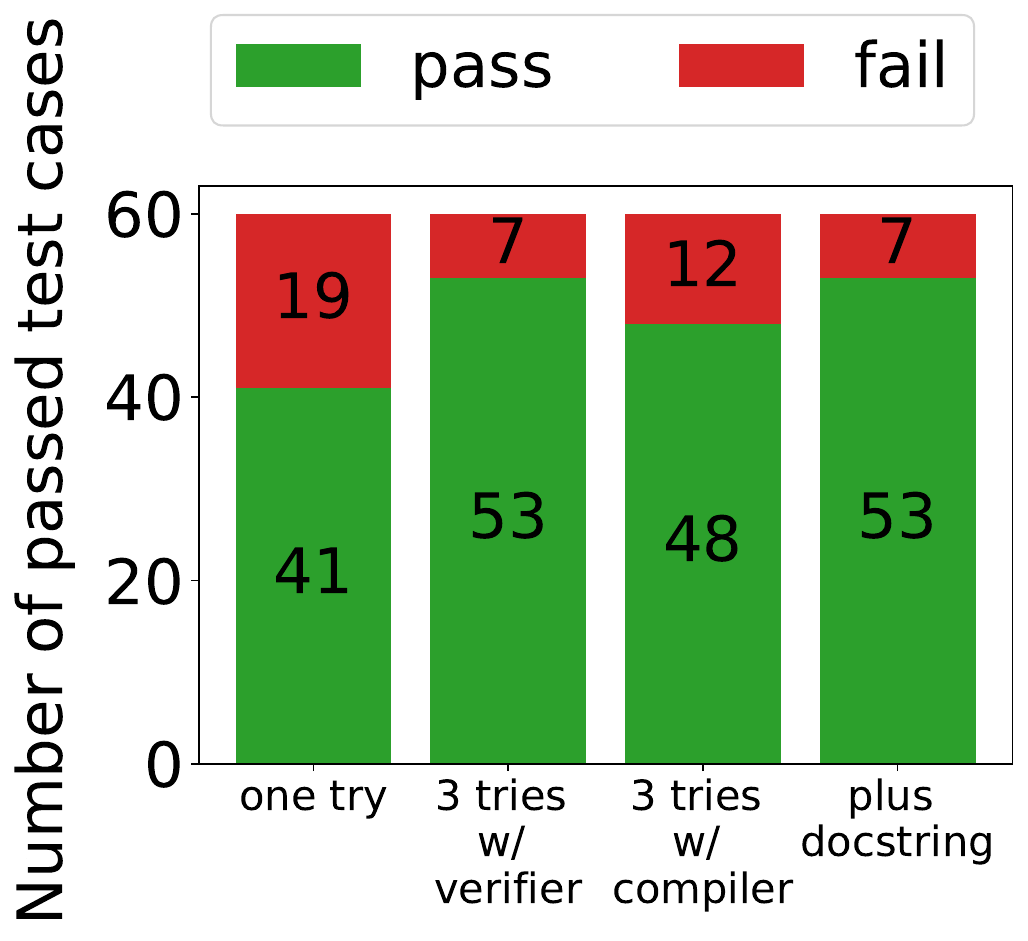}
    \caption{Code generation.}
    \label{fig:code_ablation_feedback}
\end{subfigure}
\hfill
\begin{subfigure}[b]{0.26\textwidth}
    \centering
    \includegraphics[width=1\textwidth]{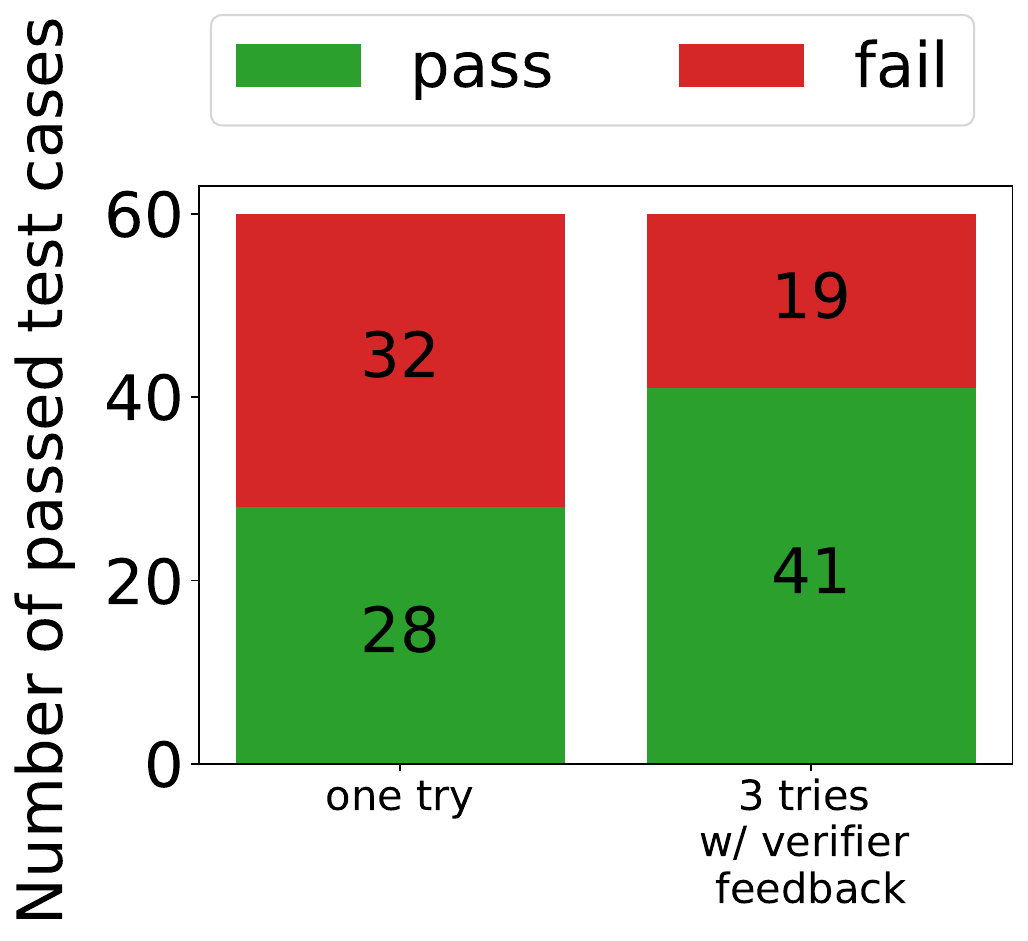}
    \caption{Annotation generation}
    \label{fig:anno_ablation_feedback}
\end{subfigure}
\hfill
\begin{subfigure}[b]{0.3\textwidth}
    \centering
    \includegraphics[width=1\textwidth]{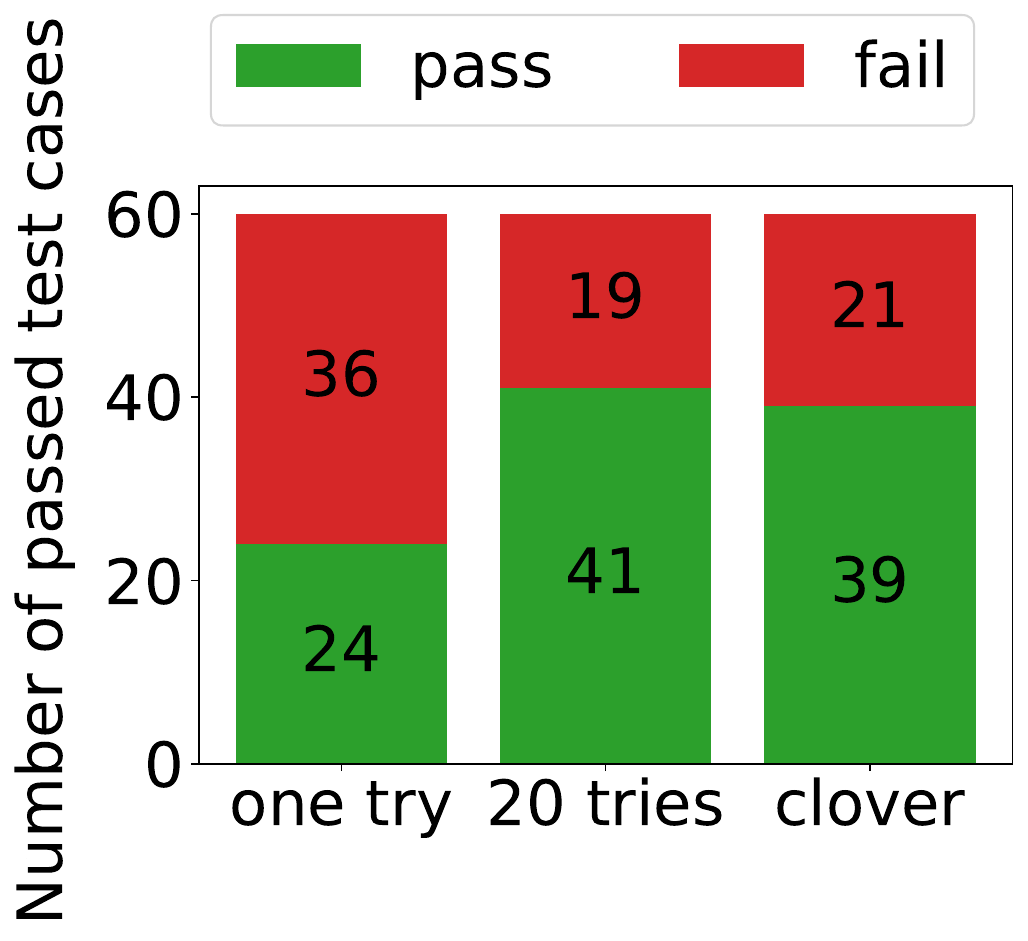}
    \caption{Code and Annotation Generation}
    \label{fig:end2end}
\end{subfigure}
\caption{Generation phase feasibility study}
\end{figure}

First, we ask \gpt to generate the code from specifications for each of the 60 examples in \dataset under various conditions. We manually checked the generated code for correctness.
Figure~\ref{fig:code_ablation_feedback} shows the results.
The first bar (``one try'') shows the result when asking \gpt to produce the code, given the annotations, in a single try. The next bar allows \gpt to try three times, each time providing the output of the \dafny compiler and verifier as feedback (See \cite[Appendix A.6]{sun2024clover} for an example of using \dafny feedback).  The next is similar but uses the output of only the \dafny compiler.  In the last bar, we allow three tries, with feedback from the \dafny compiler and verifier, and we also provide the docstring.
We see that, at its best, \gpt can correctly provide the code for 53 out of 60 examples, and it does best when it gets the most feedback from \dafny.  This suggests that \gpt is already performing reasonably well as a code synthesis tool for \dafny programs.

Second, we asked \gpt to generate full annotations (pre-conditions, post-conditions, and loop invariants) from the code alone.
Figure~\ref{fig:anno_ablation_feedback} shows the results.  In one try, \gpt succeeds on 28 of 60 programs.  Given three tries and maximal feedback from \dafny, this improves to 41 out of 60.
Though not perfect, out of the box, \gpt can produce correct annotations for the majority of programs in our simple set of benchmarks.  This suggests that using LLMs for generating annotations is feasible, and we expect that further efforts in this direction (including fine-tuning models for the task) will likely lead to even stronger capabilities.

Finally, for the last experiment, we ask \gpt to generate both the code and the annotations from the docstring alone.  Figure~\ref{fig:end2end} shows the results.  On the first try, \gpt succeeds on 24 of 60 programs.  However, if we simply do 20 independent tries and test whether \gpt succeeds on any of these tries, the number improves to 41.  This naturally raises the question: how can we leverage multiple LLM tries without having to check each one by hand?  This is exactly what the verification phase is for!  The last column in the figure shows that if we run the \sys verification phase, it accepts at least one correct answer for 39 of 41 examples for which \gpt generates a correct answer.  Further more the \sys verification check never accepts an incorrect answer.  Full results are reported in \cite[Appendix A.10]{sun2024clover}.  Thus, we can fully automatically generate 39 of 60 programs from natural language alone, with the guarantee that the generated programs pass all \sys consistency checks.  While these numbers must be improved and more complicated examples must be tried, these early results are promising and suggest that these ideas should be explored further.



 


\subsection{Verification Phase: Results on Ground-Truth Examples}
\label{sec:exp:ground_truth}

Our main experiment evaluates the capabilities of the \sys consistency checking algorithm.  During consistency checking, we consider everything that appears in the body of a method, including assertions and invariants, to be part of the code, and consider the annotations to consist only of pre- and post-conditions.  The reason for this is for modularity.  We need to be able to separate out the annotation and have it generate the code.  The assertions and invariants in the code have no context without the code, and are thus meaningless without it; moreover, the pre- and post-conditiosn contain all the information necessary to reconstruct the code.  Thus, this division makes the most sense for \sys.

For each example in \dataset, we run all 6 checks described in Section~\ref{sec:clover_schema}.  For checks that use \dafny, we use three tries and provide feedback from \dafny's compiler after each try.
We also evaluate the effect of multiple independent runs, meaning that we repeat each of the 6 checks $k$ times.  If any one of the $k$ attempts succeeds, then the check is considered to have passed.
%
The results are summarized in Table~\ref{tab:exp_summary}.
When $k=1$, we see that our \sys implementation accepts 45 of 60 correct (``ground truth'') examples and rejects all incorrect examples.
When $k=10$, \sys accepts 52 of 60 correct examples and rejects all incorrect ones.
Details on each of the 6 checks for the ground truth examples are shown in Table~\ref{tab:GroundTruthexp}.
All acceptance rates are above $80\%$.  Failures are mostly due to incorrect or imprecise reconstruction.  More details can be found in \cite[Appendix A.4.5]{sun2024clover}.
We expect that using better LLMs (either better general-purpose LLMs or LLMs fine-tuned for program verification or a specific language or both) will improve the acceptance rate.
For the complete experimental results, see \cite[Appendix A.10]{sun2024clover}.
%
%
\begin{table}[ht]
\centering
\begin{tabular}{c|cc}
\toprule
& Accept (k=1) & Accept (k=10)\\
\midrule
Ground-Truth &  $45/60$ $(75\%)$ & $52/60$ $(87\%)$ \\
Adversarial-C1 &  $0/60$ $(0\%)$ & $0/60$ $(0\%)$ \\
Adversarial-C2 &  $0/60$ $(0\%)$ & $0/60$ $(0\%)$ \\
Adversarial-C3 &  $0/60$ $(0\%)$ & $0/60$ $(0\%)$ \\
Adversarial-C6 &  $0/60$ $(0\%)$ & $0/60$ $(0\%)$ \\
\bottomrule
\multicolumn{3}{c}{$ $}\\
\end{tabular} 
\caption{Summary of the experimental results for the verification phase.}
\label{tab:exp_summary}
\end{table}
\begin{table}[]
\centering
\begin{tabular}{c|ccc}
\toprule
Ground-Truth & Accept (k=1) & Accept (k=10)\\
\midrule
anno-sound &  $60/60$ $(100\%)$ & $60/60$ $(100\%)$\\
anno-complete &  $53/60$ $(88\%)$ & $57/60$ $(95\%)$\\
doc2anno &  $51/60$ $(85\%)$ & $53/60$ $(88\%)$\\
anno2doc &  $60/60$ $(100\%)$ & $60/60$ $(100\%)$\\
code2doc &  $58/60$ $(97\%)$ & $60/60$ $(100\%)$\\ 
doc2code &  $49/60$ $(82\%)$ & $56/60$ $(93\%)$\\
\bottomrule
\multicolumn{3}{c}{$ $}\\
\end{tabular} 
\caption{Ground-truth acceptance for each of the 6 \sys checks.}\label{tab:GroundTruthexp}
\end{table}
%
%
Since our ground-truth examples are hand-written and hand-checked for correctness, it is not surprising that all pass the \dafny verifier (i.e., all annotations are sound). Annotation completeness requires successful synthesis of code from annotations, and here, we get an $88\%$ acceptance rate when $k=1$, which goes up to $95\%$ with $k=10$. The main reason for failure is incorrect generation of \dafny syntax by \gpt. In doc2anno generation, we generate annotations from docstrings. The main failure comes again from \gpt generating incorrect \dafny syntax. anno2doc and code2doc have perfect acceptance rates. On the one hand, this is because \gpt is very good at synthesizing natural language. On the other hand, our docstring equivalence checker is not very strong and skews towards acceptance. As long as they do not directly contradict each other, information omissions or additions in docstrings frequently go unnoticed by \gpt. Improving this equivalence checker is one important direction for future work. 
 doc2code generation shares the same issues as anno-complete and doc2anno: failure because of invalid \dafny syntax generation.  It also improves significantly ($93\%$ vs $82\%$) using $k=10$ instead of $k=1$.

\begin{table}[ht]
\vspace{-0.5em}
\centering
\begin{tabular}{l|ccc|p{70mm}}
\toprule
 & Code & Annotations & Docstring & Note \\ 
\midrule
C0 & - & - & - & omitted: ground-truth \\
C1 & - & - & mutated & strengthen/weaken docstring \\
C2 & - & mutated & - & weaken annotation \\
C3 & - & mutated & mutated & weaken annotations and docstring simultaneously\\
C4 & mutated & - & - & omitted: cannot pass soundness check \\
C5 & mutated & - & mutated & omitted: cannot pass soundness check \\
C6 & mutated & mutated & - & code still satisfies annotations\\
C7 & mutated & mutated & mutated & omitted: non-sense or is a variant of another ground-truth\\
\bottomrule
\multicolumn{3}{c}{$ $}\\
\end{tabular}
\caption{Categories of adversarial incorrect examples.}
\label{tab:inconsistant_category}
\end{table}

\subsection{Verification Phase: Results on MBPP-DFY-50}
\label{sec:exp:mbpp-dafny}

To explore the effectiveness of \sys on external datasets, we ran \sys on the MBPP-DFY-50 dataset~\cite{misu2024towards}, which consists of 50 \dafny programs translated by hand from Python, with docstrings and annotations. Our run revealed a number of interesting things about this dataset. First, 17 of the 50 samples are out of scope for \sys. Two are out of scope because the docstrings are not precise enough to specify a unique output for each input. The other 15 require auxiliary functions or predicates.  Extending \sys to such benchmarks is on our roadmap for future work.
Of the remaining 33 programs, 24 are accepted by \sys, and 9 are rejected. 

Looking closely at the 9 rejected samples, we determined that 6 are, in fact, incorrect:
5 have factual contradictions in their docstrings and pre-conditions; and 1 has trivial (too weak) post-conditions that do not reflect the requirements in the docstrings.\footnote{The 6 incorrect samples are shown in \cite[Appendix A.9]{sun2024clover}.}
The remaining 3 are false negatives: correct programs that do not pass all of the \sys checks. We determined that the 24 accepted benchmarks are all correct (0 false positives), once again demonstrating that \sys provides a strong filter against incorrect code. Overall, after correctly categorizing the 33 benchmarks, \sys achieves an 89\% acceptance rate ($k=10$) while maintaining a 100\% rejection rate for incorrect benchmarks.\footnote{Detailed results of the \sys checks for the 27 correct benchmarks are in \cite[Appendix A.10]{sun2024clover}).}

\subsection{Verification Phase: Results on Adversarial Examples}
\label{sec:exp:inconsistency}

As mentioned, for each program in our dataset, we created 4 adversarial incorrect versions. Here we describe them in more detail.
Table~\ref{tab:inconsistant_category} lists all possible ways we can mutate the ground-truth example, while still ensuring that it passes the \dafny verification check (anno-sound). Thus, for these examples, a naive approach using only \dafny (as in~\cite{misu2024towards}) would result in a 100\% false positive rate.  However, \sys with its 6 consistency checks is able to reject all of them (0\% false positive rate).
Category C0 is the ground-truth where we mutate nothing. Categories 1 to 7 cover all the possible ways we can mutate C0.
Category C1 contains programs in which the docstring is incorrect and the other two are the same as the ground-truth.
Category C2 contains programs in which the annotations are incorrect and the other two are the same as the ground-truth. To ensure these examples are not trivially rejected by the Dafny soundness check, we only weaken the annotations to ensure that the code still satisfies the mutated annotations.
Category C3 contains programs in which both the annotations and the docstring are mutated. The mutated annotations and docstring are simultaneously weakened, but the two are consistent.
Category C6 contains programs in which the annotations and code are consistent but inconsistent with the docstring and thus not detectable by \dafny.
Categories C4 and C5 are omitted because they are trivially rejected by the \dafny verifier (i.e., they always fail the soundness check). C7 is also omitted because it's not clear how meaningful it is to change all three, and, excluding the corner case when all three are changed to be mutually consistent, benchmarks in this category should be strictly easier to detect than those in the other categories.
\begin{table}[ht]
\vspace{-0.5em}
\centering
\resizebox{1\columnwidth}{!}{
\begin{tabular}{c|cccccccc}
\toprule
Category & \multicolumn{2}{c}{C1 Reject} & \multicolumn{2}{c}{C2 Reject} & \multicolumn{2}{c}{C3 Reject} & \multicolumn{2}{c}{C6 Reject} \\
\cmidrule(lr){2-3}\cmidrule(lr){4-5}\cmidrule(lr){6-7}\cmidrule(lr){8-9}
& k=1 & k=10 & k=1 & k=10 & k=1 & k=10 & k=1 & k=10 \\
\midrule
anno-sound &   $0/60$ ($0\%$) &  $0/60$ ($0\%$)& $0/60$ ($0\%$) & $0/60$ ($0\%$) & $0/60$ ($0\%$) &$0/60$ ($0\%$) & $0/60$ ($0\%$) &$0/60$ ($0\%$)\\
anno-complete  & $7/60$ ($12\%$) &  $3/60$ ($5\%$)& $26/60$ ($43\%$) &  $16/60$ ($27\%$)& $26/60$ ($43\%$) & $21/60$ ($35\%$)  & $33/60$ ($55\%$) &$27/60$ ($45\%$) \\
doc2anno & $57/60$ ($95\%$) &  $54/60$ ($90\%$)& $60/60$ ($100\%$) & $60/60$ ($100\%$) & $44/60$ ($73\%$) & $30/60$ ($50\%$) & $60/60$ ($100\%$) &$60/60$ ($100\%$) \\
anno2doc &  $42/60$ ($70\%$) &  $34/60$ ($57\%$)& $24/60$ ($60\%$) &  $13/60$ ($22\%$)& $24/60$ ($60\%$) & $4/60$ ($7\%$) & $42/60$ ($70\%$) &$27/60$ ($45\%$)\\ 
code2doc &  $57/60$ ($95\%$) &  $54/60$ ($90\%$)& $0/60$ ($0\%$) & $0/60$ ($0\%$) & $51/60$ ($85\%$) & $43/60$ ($72\%$) & $43/60$ ($72\%$) &$40/60$ ($67\%$)\\
doc2code & $39/60$ ($65\%$) &  $37/60$ ($62\%$)&  $11/60$ ($18\%$) &  $4/60$ ($7\%$)& $31/60$ ($52\%$) &$18/60$ ($30\%$)  & $58/60$ ($97\%$) &$55/60$ ($92\%$) \\ 
\bottomrule
\multicolumn{9}{c}{$ $}\\
\end{tabular}}
\caption{Rejection rates for adversarial incorrect examples.}
\label{tab:inconsistent_exp}
\vspace{-0.5em}
\end{table}

Table~\ref{tab:inconsistent_exp} shows the results of the 6 checks for each category.  We observe that doc2anno has the highest rejection rate. This is because we use \dafny to do a formal equivalence check, which guarantees that only logically equivalent annotations are accepted.
Overall, there are no false positives (no incorrect example passes all 6 checks), as summarized in Table~\ref{tab:exp_summary}.\footnote{For complete results, see Tables in \cite[Appendix A.10]{sun2024clover}.}


\subsection{A Preliminary Study with Verus}
As mentioned, we chose \dafny for our primary study because of its maturity as a deductive verification tool.  A natural question is how \sys performs with other systems and languages.  To gain some understanding of this, we did a preliminary study using Verus~\cite{lattuada2023verus}, a deductive verification tool for a subset of the Rust programming language.
Verus and \dafny share the common goal of integrating verification into the development process, but they differ in several ways. For instance, Verus is designed to be more performant but less automatic than \dafny.  This means that it often requires more proof effort than \dafny to verify the same program.  Verus is also less mature than \dafny, having been developed only recently.

We implemented 41 ground-truth examples in Verus~\cite{lattuada2023verus} and 
used the same approach used with \dafny to perform the \sys consistency checks (except that formal checks were done with the Verus tool instead of \dafny).  Also, because the Verus specification format is very new, we started each LLM prompt with a few simple examples of Verus specification syntax.  For our 41 examples, \sys accepts 32 of 41 when $k=1$ and 36 out of 41 when $k=10$.  Full results are shown in Table~\ref{tab:verus-gt}.
These early results suggest that \sys can be used with other languages and deductive verification tools.

\begin{table}[]
\centering
\begin{tabular}{c|ccc}
\toprule
Ground-Truth & Accept (k=1) & Accept (k=10)\\
\midrule
anno-sound &  $41/41$ $(100\%)$ & $41/41$ $(100\%)$\\
anno-complete &  $39/41$ $(95\%)$ & $40/41$ $(98\%)$\\
doc2anno &  $33/41$ $(80\%)$ & $36/41$ $(88\%)$\\
anno2doc &  $41/41$ $(100\%)$ & $41/41$ $(100\%)$\\
code2doc &  $41/41$ $(100\%)$ & $41/41$ $(100\%)$\\ 
doc2code &  $41/41$ $(100\%)$ & $41/41$ $(100\%)$\\
\bottomrule
\multicolumn{3}{c}{$ $}\\
\end{tabular} 
\caption{Verus Ground-truth acceptance for each of the 6 \sys checks.}
\label{tab:verus-gt}
\end{table}

\section{Related Work}


\paragraph{Code Generation}
Besides well-known work \cite{DBLP:journals/corr/abs-2107-03374,DBLP:conf/iclr/ChengX0LNHXROZS23,li2022competition} on code generation using LLMs,
\cite{DBLP:journals/ftpl/GulwaniPS17} is a survey on program synthesis before the era of LLMs.
Other works using neural approaches for program synthesis include \cite{DBLP:journals/corr/abs-2108-07732,DBLP:journals/pacmpl/BowersOWGTES23,DBLP:conf/acl/YinN17}.
To scale up code generation, researchers have tried to decompose the whole task into smaller steps 
\cite{zelikman2023parsel,DBLP:conf/pldi/EllisWNSMHCST21,DBLP:journals/pacmpl/BowersOWGTES23} and to use execution traces \cite{ding2023traced,DBLP:conf/iclr/ShiDES22}.
While the aforementioned works synthesize code from natural language, another common theme is to synthesize programs from specifications \cite{DBLP:conf/fmcad/AlurBJMRSSSTU13,DBLP:journals/ftpl/ChaudhuriEPSSY21,DBLP:conf/pldi/PolikarpovaKS16,solar2008program}.
Translation between natural and formal language has also been studied in
\cite{DBLP:conf/lrec/GhoshSMSSZB22,DBLP:journals/corr/abs-2206-01962,sun2023towards}, and 
LLMs have been used to predict program invariants~\cite{liu2023towards,DBLP:conf/icml/PeiBSSY23,wu2023lemur}.

Various approaches have been explored for self-correction in code generation, as surveyed in \cite{pan2023automatically}, including self-consistency~\cite{wang2022self}, self-debugging~\cite{chen2023teaching,saunders2022self}, and self-improvement~\cite{madaan2024self}.
In \cite{olausson2023self}, self-debugging has shown to be limited compared to human-level debugging.

\paragraph{Verified Generation}
Prior works acknowledge that verifying whether a generated program is correct is challenging.
In \cite{DBLP:journals/corr/abs-2305-01210}, a test-case-based approach 
is demonstrated to be insufficient.
Other previous attempts include \cite{DBLP:journals/corr/abs-2210-00848}, which asks the model to generate assertions along with the code,
and \cite{DBLP:conf/iclr/ChenZNZLLC23,DBLP:journals/corr/abs-2107-03374,ryan2024code,chen2022codet},
which study the generation of unit tests and how to use the generated tests to increase the generation accuracy.
There is also a line of work \cite{DBLP:journals/corr/abs-2110-14168,DBLP:conf/nips/InalaWYCELMG22,DBLP:journals/corr/abs-2206-02336,DBLP:conf/icml/ZhangYHLYF023} on a learning-based approach for verifying correctness. \cite{DBLP:conf/nips/InalaWYCELMG22,li2022competition,DBLP:conf/emnlp/ShiFGZW22,DBLP:conf/nips/Wei0SBIXCLZ22}
study various approaches for reranking a model's output, and
\cite{DBLP:journals/corr/abs-2305-14752} propose a self-repair method combining LLMs and bounded model checking to locate software vulnerabilities and derive counterexamples.

Finally, there has recently been a marked and rapid surge of interest in using LLMs to generate formal annotations for verification purposes.
\cite{wen2024enchanting} generates specifications by leveraging LLMs and techniques from static analysis and program verification.
Research in specific domains includes examples like \cite{mohammed2024enabling}, which proposes a framework for porting C to Checked-C to enable memory safety for C programs, and \cite{mondal2023llms}, which uses LLMs to synthesize verified router configurations in networking. 
Most closely related to our work is \cite{brandfonbrener2024verified}, which uses Monte Carlo Tree Search to help with the multi-step synthesis of annotated Dafny programs,
and \cite{misu2024towards}, which explores prompting techniques for generating Dafny programs.
In contrast to our work, both of these focus on generation rather than verification. Furthermore, they use only the soundness check, whereas \sys requires a stronger set of six consistency checks.

\section{Conclusion}
We have introduced \sys, a paradigm for closed-loop verifiable code generation, together with a new dataset \dataset featuring 60 hand-crafted \dafny examples.
We reduce the problem of checking correctness to the more accessible problem of checking consistency.
Initial experiments using \gpt on \dataset are promising. We show an $87\%$ acceptance rate for ground-truth examples in \dataset and a $100\%$ rejection rate for incorrect examples.
\sys also accurately detects 6 incorrect samples and accepts $89\%$ correct ones in the existing human-written dataset MBPP-DFY-50~\cite{misu2024towards}.
There are many avenues for future work, including: better verification tools, improving LLM capabilities for generating code, annotations, and docstrings, improving  LLM capabilities for understanding \dafny and Verus syntax, and scaling up to more challenging examples.

\subsection*{Acknowledgements}
This work was supported in part by an Amazon Research Award and the Stanford Center for Automated Reasoning (Centaur).

\newpage


\printbibliography
\newpage
\appendix
\section{Appendix}
\setlength{\columnseprule}{1pt}
\def\columnseprulecolor{\color{black}}

\subsection{An Analytical Model for \sys Reconstruction Tests}
\label{sec:clover_analytical}

As described in Section~\ref{sec:clover:consistency-check}, all but one of the six Clover consistency checks relies on reconstructing one of the components (see Figure~\ref{fig:overview}).  These reconstructions rely on assumptions about the LLM model used for reconstruction that have, until now, been implicit.  In this section, we make these assumptions explicit and provide a theoretical model and analysis for those assumptions. 
For the purpose of the analysis we focus on a single directed edge from domain $A$ to domain $B$ (e.g., code to docstring).

Assume each domain is equipped with a semantic equivalence relation, denoted by $\equiv$.
Each domain can therefore be partitioned into equivalence classes.
For $X\in\{A,B\}$, we use $e(X)$ to denote the set of equivalence classes of $X$, and for $x \in X$ we use $[x]$ to denote the equivalence class $x$ belongs to.
For docstrings, the equivalence relation represents semantic equivalence as understood by a human expert; for code, the equivalence relation is functional equivalence; and for annotations, it is logical equivalence.

We further assume a \emph{ground truth consistency relation} between $A$ and $B$, denoted by $G \subseteq A \times B$. The ground truth consistency represents the consistency we assume to exist between docstrings, annotations, and code, as described in Section~\ref{sec:clover_schema}.
We assume the consistency relation satisfies the following properties that link it to the equivalence relation:
For any $x, x'\in A$ and $y, y' \in B$,
$(x \equiv x' \wedge y \equiv y') \rightarrow ((x, y) \in G \leftrightarrow (x', y')\in G)$ and
$((x, y) \in G \wedge (x', y')\in G) \rightarrow (x \equiv x' \leftrightarrow y \equiv y')$.
That is, consistency is preserved when substituting equivalent objects, and any object may be consistent with at most one equivalence class from the other domain.

We now formally define and analyze the \emph{single-edge \sys consistency test}, which aims to be an approximate test for $G$. For the analysis, we assume a probability distribution $\mathcal{D}$ on $A \times B$. The test relies on a \emph{transfer model} and the analysis assumes it is \emph{transfer-rational}, as defined below.

\begin{definition}[Transfer Model]
Given domains $A$ and $B$, a transfer model for $A$ and $B$ is a function $M: A\times B \rightarrow \mathbb{R}$
such that for each $x\in A$, $M(x, \cdot)$ is a probability distribution over $B$.
Here $M(x, y)$ denotes the probability of transferring $x\in A$ to $y \in B$.
\end{definition}

\newcommand{\argmax}[1]{\underset{#1}{\operatorname{arg}\,\operatorname{max}}\;}

\begin{definition}[Transfer-Rational Model]
Let $M$ be a transfer model for $A$ and $B$.
We say $M$ is \textit{transfer-rational} if for each $x\in A$ there is a unique $[y]\in e(B)$ that maximizes $\sum_{y'\in [y]} M(x, y')$.
In this case, we define the \textit{transfer function of} $M$,  $f^M: A \rightarrow e(B) = \lambda x. \, \argmax{[y]\in e(B)} \sum_{y'\in [y]} M(x, y')$.

\end{definition}

Intuitively, the transfer model is meant to approximate a mapping based on the ground truth consistency ($G$).
In the context of \sys, the domains are among docstring, annotation, and code, and the transfer model is given by an LLM (\gpt). For example, when $A$ is docstrings and $B$ is annotations, the distribution $M(x, \cdot)$ represents the output distribution of \gpt on an input docstring $x$ with a suitable prompt for generating an annotation corresponding to the docstring $x$.
In our evaluation, we use 3 tries with \dafny feedback to run the reconstruction test.
In this case, the transfer model is given by this combined use of \gpt and \dafny.

We now fix a transfer-rational model $M$, and define the single-edge \sys consistency check.

\begin{definition}[Single-Edge \sys Consistency Check]
For input $x\in A, y\in B$, the single-edge \sys consistency check (for the edge from $A$ to $B$) is a procedure that draws $y'$ from the distribution $M(x,\cdot)$, and then accepts if $y' \equiv y$ and otherwise rejects.
\end{definition}

Note that the check relies on being able to check equivalence in domain $B$.\footnote{We assume a perfect equivalence check to keep the analysis simple and illustrative. In practice, the equivalence tests do incur some imprecision. But accounting for this imprecision using a probabilistic model is cumbersome because the distribution on the equivalence checks \sys performs depends on both the input distribution and on the transfer model.}

We now analyze the probability that the single-edge \sys consistency check is correct. Our analysis relies on two assumptions: one relating the transfer model $M$ with the ground truth consistency $G$, and another ensuring that $M$'s distributions are concentrated.

\begin{assumption}[Consistency Alignment]
\label{asmp:gtconsistency}
Let $c_1$ be the probability that $y\in f^M(x)$
when $x, y$ are sampled from $A\times B$ according to $\mathcal{D}$ conditioned on $(x, y)\in G$.
Similarly, let $c_0$ be the probability that $y\in f^M(x)$
when $x, y$ are sampled from $A\times B$ according to $\mathcal{D}$ conditioned on $(x, y)\not\in G$.
We assume that $c_1$ is close to 1, and $c_0$ is close to 0.
\end{assumption}

\begin{assumption}[Concentration]
\label{asmp:transfer-function}
Consider $x, y$ sampled from $A\times B$ according to $\mathcal{D}$
conditioned on $(x, y) \in G$ and $y\in f^M(x)$.
We assume that for some significant $0<l\leq 1$ (e.g., 30\%), the following holds with probability $\geq p_c$ ($p_c$ close to 1): 
$\sum_{y'\in f^M(x)} M(x, y') \geq l$.
Similarly, consider $x, y$ sampled from $A\times B$ according to $\mathcal{D}$
conditioned on $(x, y) \notin G$ and $y\notin f^M(x)$.
We assume that for some negligible $u$, the following holds with probability $\geq p_c$:
$\max_{[y_1]\in e(Y), [y_1]\neq f^M(x)} \sum_{y_2\in [y_1]} M(x, y_2)\leq u$.
\end{assumption}

Intuitively, the concentration assumption means that with high probability ($\geq p_c$), sampling from $M$ is the same as applying $f^M$, and specifically that the second most likely equivalence class is much less likely than the maximal one (i.e., the one given by $f^M$).

\begin{theorem}
\label{thm:clover}
Under Assumptions~\ref{asmp:gtconsistency} (Consistency) and ~\ref{asmp:transfer-function} (Concentration), 
consider $(x, y)$ sampled from $A\times B$ according to $\mathcal{D}$
conditioned on $(x, y)\in G$;
the single-edge \sys consistency check will accept $(x, y)$ with probability 
$A\geq l \cdot p_c \cdot c_1$.
Similarly, consider $(x, y)$ sampled from $A\times B$ according to $\mathcal{D}$ 
conditioned on $(x, y)\notin G$;
the single-edge \sys consistency check will 
will accept with probability $R \leq u \cdot p_c\cdot (1 - c_0) + (1-p_c)(1-c_0) + c_0$.
\end{theorem}

The proof of Theorem~\ref{thm:clover} is in Appendix~\ref{sec:appendix:proof}.

Theorem~\ref{thm:clover} ensures that under our assumptions, the probability of accepting a consistent input is significant, and the probability of accepting an inconsistent input is negligible. We can increase the gap by repeating the reconstruction test several times and accepting if any of them accept. 
As discussed in Section~\ref{sec:evaluation}, our evaluation shows the results for both 1 and 10 reconstruction attempts.

\textbf{From single-edge to full \sys consistency checking.} Our analysis focused on a single, directed reconstruction edge from Figure~\ref{fig:overview}, while full \sys consistency checking uses five reconstruction edges and a single verification edge, and accepts only if all six checks accept. We do not attempt to theoretically analyze the full check, because we do not assume the edges to be independent (so multiplying acceptance probabilities is not necessarily meaningful). 
In our experiments, we empirically measure the acceptance rate of each edge, and also observe that the edges are not independent (see Section~\ref{sec:instantiation}).
In real experiments, the combined use of \gpt and \dafny may not satisfy our assumptions because of the tools' limitations (\dafny may time out or return unknown, and \gpt may make mistakes or hallucinate). Especially the $u$ in Assumption~\ref{asmp:transfer-function} could be non-negligible.
However, the end-to-end evaluation shows that the six checks together do give promising true positive and false positive rates.
The analytical model can be treated as one guide to understanding what properties of the reconstruction model are helpful for ensuring accurate reconstruction results.

\subsection{Explaining the Evaluation}
\label{sec:instantiation}

\begin{table}[ht]
\footnotesize
\centering
\begin{tabular}{c|ccccc}
\toprule
edge & \makecell{Accept \\Correct \\$A$ }& \makecell{Accept \\Incorrect \\$R^{C1}$} & \makecell{Accept\\ Incorrect\\ $R^{C2}$} & \makecell{Accept \\Incorrect \\$R^{C3}$} &
\makecell{Accept \\Incorrect \\$R^{C6}$}\\

\midrule
anno-sound &  $1$    & $-$ & $-$ & $-$ & $-$\\
anno-complete  &  $0.88$   & $-$& $-$ & $-$ & $-$\\
doc2anno &  $0.85$    & $0.05$ & $0$ & $-$ & $0$\\
anno2doc &  $1$    & $0.30$ & $0.4$ & $-$ & $0.30$\\
code2doc &  $0.97$    & $0.05$ & $-$ & $0.15$ & $0.28$\\ 
doc2code  &  $0.82$    & $-$ & $-$ & $0.48$ & $0.03$\\
\bottomrule
\end{tabular} 
\caption{Empirically measured values for $A$ and $R$ when $k=1$. Entries shown as ``$-$'' are omitted because for that category and check, the assumptions of the analytical model are violated.}
\label{tab:plugin}
\end{table}

Here, we empirically estimate the values of $A$ and $R$ from Theorem~\ref{thm:clover} based on our experiments.
That is, we estimate the acceptance rate for correct and incorrect inputs for each directed edge.
Each cell in 
Table~\ref{tab:plugin} represents the percentage of reconstructed components that successfully pass the equivalence check in the five categories: ground-truth
, C1, C2, C3 and C6 (Table~\ref{tab:inconsistant_category}).%
\footnote{Note that our incorrect examples are constructed with the aim of making them hard to reject, i.e., by considering only the cases that can pass Dafny verification. The measured values for $R$ are thus likely to be higher than the value for a more natural distribution.}

As mentioned above, in the first column, the discrepancy between the measured acceptance rate and the ideal perfect acceptance rate comes partly from reconstruction failures and partly from equivalence checker failures. For example, the doc2anno acceptance rate is $0.85$, not $1$. Apart from the failure to generate the correct annotation, there are also cases where the generated annotation is correct but unable to be verified by our annotation equivalence checking template in Section~\ref{sec:dataset} (See Appendix~\ref{sec:sup:anno_template} for an example).

Overall, the measured aggregated acceptance rate for the first column is $0.75$.  This is higher than would be expected if each check were independent (the product of the entire column is $0.59$).  This is because, in practice, they are not independent: easier examples that pass the tests on one edge tend to also pass the tests on other edges. In C2 and C6, doc2anno has a zero acceptance rate, and the overall acceptance is zero. In C1 and C3, none of the edges are zero, but the overall acceptance is still zero. 
Note that the anno2doc and code2doc acceptance rate is high for C1, C2, and C6. 
This is because our current docstring equivalence checker is good at detecting contradictory information but not the addition or omission of information due to a slightly strengthened or weakened annotation.

\subsection{Proof of Theorem~\ref{thm:clover}}
\label{sec:appendix:proof}
\begin{itemize}
    \item[1.] Let $(x, y)$ be sampled from $\mathcal{D}$ with the condition $(x, y)\in G$.
    From Assumption~\ref{asmp:gtconsistency}, with probability $\geq c_1$, we have $y\in f^M(x)$.
    Then, according to Assumption~\ref{asmp:transfer-function} and the perfect equivalence oracle, with probability $p_c$, the reconstruction from $x$ to $y$ will succeed with probability $\geq l$.
    Therefore, the accept probability is $\geq l \cdot p_c\cdot c_1$, denoted as $A$.
    \item[2.] Let $(x, y)$ be sampled from $\mathcal{D}$ with the condition $(x, y)\notin G$. There are 3 cases:
    \begin{itemize}
        \item From Assumption~\ref{asmp:gtconsistency}, with probability $c_0$, there is $y\in f^M(x)$, and it is trivial that the accept probability $\leq 1$.
        \item With probability $1 - c_0$, there is $y\notin f^M(x)$.
        Then from Assumption~\ref{asmp:transfer-function}, with probability $p_c$, the reconstruction from $x$ to $y$ will succeed with probability $\leq u$.
        \item Finally, the last case is that the bounds in Assumption~\ref{asmp:transfer-function} do not hold, which will happen with probability $(1-c_0)(1 - p_c)$.  Clearly, in this case, the accept probability $\leq 1$.
    \end{itemize}
    By aggregating all the cases, the accept probability is $\leq c_0 + (1-c_0)\cdot p_c\cdot u + (1-c_0)(1-p_c)$.
\end{itemize}

\subsection{Discussion}
\label{sec:discussions}
\subsubsection{Limitations}
\label{sec:limitations}
There are many limitations in the proposed paradigm.  For one, the capabilities of LLMs (\gpt in particular) are limited.
The generation of docstrings, annotations, and code also has inherent limitations.
For example, our use of annotations is only for specifying functionality, not implementation details,
e.g., an annotation can force an array to be sorted but cannot easily restrict the algorithm used for sorting.
In this paper, as a first step, we only aim to check functional consistency (correctness), not the performance of the implementation. 

As mentioned in Section~\ref{sec:clover},
if the oracle used for consistency checking is misaligned with human understanding (ground-truth), e.g., if it interprets a sorting algorithm as getting the maximum value, there is no way to correct it without human intervention.
But in practice, we think this will rarely happen (see Assumption~\ref{asmp:gtconsistency}).
As another example, if the docstring, annotations, and code all miss the same edge case, the error cannot be detected. While such an example is internally consistent, it may not be consistent with human understanding of good coding practice. Since a LLM is trained on a vast corpus of human-written data, it is inherently designed to align with human understanding. This misalignment occurs so infrequently that we have opted not to include it in the main paper. To date, we have not detected this issue in our experiments. 
To achieve our eventual vision for \sys,  we expect that additional breakthroughs, or additional human-in-the-loop steps, or both, may be needed.

\subsubsection{\sys Variants}
\label{sec:variants}
\sys checks three components for consistency.  However, other variants are possible.
Currently, most attempts at code generation produce only the code and docstring.
We expect that a \sys-like approach with only code and docstrings would help detect some inconsistencies, but would not ensure implementation correctness, as docstrings are not sufficiently precise.
Incorporating unit tests into \sys is a potential improvement we've earmarked for future endeavors. We recognize the potential advantages of unit tests; however, they come with their own set of limitations. Admittedly, in certain scenarios, unit tests can provide a quick and effective sanity check on system functionalities. However, generating unit tests can sometimes prove more complex than creating annotations.
Unit tests, if not transparent, can be difficult or even impossible to explain, eroding user confidence due to their opacity.
If an LLM is adept at producing effective unit tests, it suggests an ability to anticipate execution outcomes to a certain degree. However, full-fledged execution with numerous computational steps remains an unsolved challenge for LLMs.
Additionally, compared to annotations, unit tests offer a less robust assurance of system correctness.
\subsubsection{Future Research}
\label{sec:future_work}

A successful \sys paradigm relies on many components.
To maximize the capabilities of \sys, there are several foundational topics that should be explored.  Each of these areas can be advanced individually, and notably, they possess wider applicability beyond just the scope of \sys.

One foundational element is the ability to generate high-quality code, annotations, and docstrings. Clearly, the verification phase cannot compensate for poor generation, it can only detect and flag such examples.  Better equivalence checking would also improve \sys's abilities. 
Currently, it is most challenging to perform equivalence checks on docstrings.  Equivalence of annotations relies on the logical power of solvers in the back end of Dafny, whose performance and capabilities can be improved. Equivalence checking for code is also challenging; techniques like fuzzing and concolic testing (and even full formal equivalence checking) could be leveraged to improve this step.

\subsubsection{Data Contamination}
\label{sec:data_contamination}
We want to point out that the current version of \dataset has some limitations.
We hand-crafted it starting with simple textbook-level examples so as to have a baseline for more advanced work.
But we must acknowledge the possibility of indirect data contamination.
While we expect most of our examples are not explicitly present in the training data (Dafny is not a widely-used language, and we wrote the examples ourselves), there's a considerable chance that \gpt has encountered analogous data in the past.
Even if only code with a similar functionality in another language has been seen in the training data, our hand-crafted examples can be affected.
Some soft evidence for this is the observation that \gpt can sometimes generate the correct code even with incomplete docstrings or annotations.
We noticed that often, a descriptive function signature alone can be quite revealing. To mitigate this potential bias in our experiments, we opted to replace the function names with generic, non-descriptive identifiers.
In future work, we plan to update \dataset with more sophisticated examples, which we hope will help mitigate the risk of inaccurate conclusions due to data contamination in future experiments.

\subsubsection{Reasons for Reconstruction Failure using GPT-4}
\label{sec:failure_reason}
We have observed that \gpt is not very capable at producing correct syntax for the latest version ($4.0.0$) of \dafny, likely due to limited training data. One can imagine that an LLM trained or fine-tuned on \dafny $4.0.0$ would easily acheive a higher acceptance rate. Some evidence that \gpt is not familiar with the current \dafny syntax is as follows. Annotations must include a \verb|modifies clause| or \verb|reads clause| for methods that access memory. In particular, \verb|reads array| is required when the method reads from \verb|array|, and \gpt misses it almost $100\%$ of the time on its first try at generation. Luckily, with \dafny's compiler-generated error messages, \gpt is often able to add the needed \verb|modifies clause| or \verb|reads clause|. Another example is that \dafny used to require annotations to be separated by semicolons, or assert explicitly that an array is not null \verb|requires array!=null| in the pre-conditions.  These are not required any more, but \gpt still largely adheres to those deprecated rules.

\subsection{Open Model Results}
\label{sec:finetune}

We present the results of CodeLlama-34b on Clover tests in Table~\ref{table:codellama}. The results indicate that CodeLlama-34b is incapable of dealing with Dafny code, neither verification logic. In the early stages of this project, we briefly experimented with several other open models and discovered that most are significantly lacking in their ability to handle tasks related to program verification. While many open models have shown impressive results in other popular datasets, our dataset offers a perspective on the extent of the gap in knowledge coverage between open models and GPT-4. Specifically, open models perform notably poorly with low-resource languages. But GPT-4 highlights the potential of AI in verification tasks. Therefore, one of our goals is to showcase its promise and to garner more attention towards integrating verification into the workflow from both AI and verification specialists.

\begin{table}[ht!]
\centering
\begin{tabular}{lcccc}
\toprule
\textbf{Metric} & \textbf{anno\_sound} & \textbf{anno\_complete} & \textbf{code2doc} & \textbf{Clover 6 edges} \\
\midrule
ground truth & 60/60 & 6/60 & 8/60 & 2/60 \\
\toprule
\textbf{Metric} & \textbf{doc2code} & \textbf{anno2doc} & \textbf{doc2anno} &  \\
\midrule
ground truth & 2/60 & 48/60 & 2/60 &\\
\bottomrule
\end{tabular}
\caption{Ground Truth Acceptance by CodeLlama-34b}
\label{table:codellama}
\end{table}
\subsection{Dafny Feedback Example}
\label{sec:appendix:dafny_feedback}
In this example, the first try at generating annotations (Listing~\ref{lst:updategen}) fails given the input in Listing~\ref{lst:UpdateElements}. But after receiving the error message in Listing~\ref{lst:dafnyfeedback}, \gpt is able to add the necessary annotations \verb|modifies a|.

\begin{minipage}{0.48\textwidth}
\begin{lstlisting}[style=dafnystyle,basicstyle=\ttfamily\scriptsize,label=lst:UpdateElements,caption={UpdateElements Example Input}]
method foo(a: array<int>)
  //TOFILL
{
  a[4], a[8] := a[4] + 3, a[8] + 1;
  a[7], a[8] := 516, a[8] - 1;
}
\end{lstlisting}
\end{minipage}
\hfill
\begin{minipage}{0.48\textwidth}
\begin{lstlisting}[style=dafnystyle,basicstyle=\ttfamily\scriptsize,label=lst:updategen,caption={Generated Annotations for UpdateElements}]
requires a != null && a.Length > 8
ensures a[4] == old(a[4]) + 3
ensures a[7] == 516
ensures a[8] == old(a[8])
\end{lstlisting}

\begin{lstlisting}[style=terminal,label=lst:dafnyfeedback,caption={Dafny Feedback}]
Error: assignment might update an array element not in the enclosing context's modifies clause
\end{lstlisting}
\end{minipage}

\subsection{Supplimentary Template and Examples}
\label{sec:sup:anno_template}
In Listing~\ref{anno_template}, we give a template for verifying annotation equivalence for the ground-truth example \verb|max_array| (Listing~\ref{maxArray}).
Annotation equivalence checking is done by verifying the template with \dafny's verifier. If the lemmas \texttt{pre\_eq} and \texttt{post\_eq} are both verified, then it means that \dafny has successfully verified the equivalence of pre- and postconditions respectively.

In more detail, \verb|predicate pre_original| states the full preconditions of the ground-truth example, and \verb|predicate post_original| states the full postconditions. \verb|predicate pre_gen|'s body will be replaced by the generated preconditions and \verb|predicate post_gen|'s body will be replaced by the generated postconditions.
The lemma \verb|pre_eq| states that the generated preconditions are true if and only if the original preconditions are true. The lemma \verb|post_eq| states that the generated postconditions are true if and only if the original postconditions are true.
The above example is simple enough to be proven by Dafny's verifier.

Note that the template is sound but not complete, that is, there could be cases when two predicates are indeed equivalent but \dafny cannot prove it. 
An example is shown in Listing~\ref{only_once_filled_template}.

\begin{lstlisting}[style=dafnystyle, caption=maxArray,label=maxArray]
method maxArray(a: array<int>) returns (m: int)
  requires a.Length >= 1
  ensures forall k :: 0 <= k < a.Length ==> m >= a[k]
  ensures exists k :: 0 <= k < a.Length && m == a[k]
{
  m := a[0];
  var index := 1;
  while (index < a.Length)
    invariant 0 <= index <= a.Length
    invariant forall k :: 0 <= k < index ==> m >= a[k];
    invariant exists k :: 0 <= k < index && m == a[k];
    decreases a.Length - index
  {
    m := if m>a[index] then  m else a[index];
    index := index + 1;
  }
}
\end{lstlisting}

\begin{lstlisting}[style=dafnystyle,  caption=Annotation Equivalence Checking Template for maxArray,label=anno_template]
predicate pre_original(a: array<int>,m: int)
  reads a
{
  ( a.Length >= 1)
}

predicate pre_gen(a: array<int>,m: int)
  reads a
{
  true // (#PRE) && ... (#PRE)
}

lemma pre_eq(a: array<int>,m: int)
  ensures pre_original(a,m ) <==> pre_gen(a,m )
{
}

predicate post_original(a: array<int>,m: int)
  requires pre_original(a,m)
  reads a
{
  ( forall k :: 0 <= k < a.Length ==> m >= a[k]) &&
  ( exists k :: 0 <= k < a.Length && m == a[k])
}

predicate post_gen(a: array<int>,m: int)
  requires pre_original(a,m)
  reads a
{
  true // (#POST) && ... (#POST)
}

lemma post_eq(a: array<int>,m: int)
  requires pre_original(a,m )
  requires pre_gen(a,m )
  ensures post_original(a,m ) <==> post_gen(a,m )
{
}
\end{lstlisting}

\begin{lstlisting}[style=dafnystyle,  caption=Instantiated Annotation Equivalence Checking Template for only\_once. The original and generated postconditions describe the same property: element key only appears once in the array a. But they cannot be verified as equivalent by the annotation template. Lemma post\_eq will fail with an empty body.,label=only_once_filled_template]
predicate pre_original<T(==)>(a: array<T>,key: T,b:bool)
  reads a
{
  true
}

predicate pre_gen<T(==)>(a: array<T>,key: T,b:bool)
  reads a
{
  true
}

lemma pre_eq<T(==)>(a: array<T>,key: T,b:bool)
  ensures pre_original(a,key,b ) <==> pre_gen(a,key,b )
{
}

predicate post_original<T(==)>(a: array<T>,key: T,b:bool)
  requires pre_original(a,key,b)
  reads a
{
  ( (multiset(a[..])[key] ==1 ) <==> b)
}

predicate post_gen<T(==)>(a: array<T>,key: T,b:bool)
  requires pre_original(a,key,b)
  reads a
{
  (b <==> ((exists i :: 0 <= i < a.Length && a[i] == key) && (forall i, j :: 0 <= i < j < a.Length && a[i] == key ==> a[j] != key)))
}

lemma post_eq<T(==)>(a: array<T>,key: T,b:bool)
  requires pre_original(a,key,b )
  requires pre_gen(a,key,b )
  ensures post_original(a,key,b ) <==> post_gen(a,key,b )
{
}
\end{lstlisting}


\subsection{Input/Output Tests Template in \dataset}
\label{sec:sup:unit_tests}
Here is an example of the input/output test code we use in \dataset for code equivalence check. We compare the output from Listing~\ref{maxArrayTests} with the output when the \verb|method maxArray| implementation is replaced by the generated code. If the outputs are equal, we consider the two codes to be equivalent.
\begin{lstlisting}[style=dafnystyle,  caption=Ground truth unit tests for max\_array,label=maxArrayTests]
method maxArray(a: array<int>) returns (m: int)
  requires a.Length >= 1
  ensures forall k :: 0 <= k < a.Length ==> m >= a[k]
  ensures exists k :: 0 <= k < a.Length && m == a[k]
{
  m := a[0];
  var index := 1;
  while (index < a.Length)
    invariant 0 <= index <= a.Length
    invariant forall k :: 0 <= k < index ==> m >= a[k]
    invariant exists k :: 0 <= k < index && m == a[k]
    decreases a.Length - index
  {
    m := if m>a[index] then  m else a[index];
    index := index + 1;
  }
}

method TestMethod(){
  var a1 := new int[5];
  a1[0] := 1; a1[1] := 2; a1[2] := 3; a1[3] := 4; a1[4] := 5;
  var test1 := maxArray(a1);
  print("Test 1: maxArray([1,2,3,4,5]) = ", test1, "\n");

  var a2 := new int[5];
  a2[0] := -1; a2[1] := -2; a2[2] := -3; a2[3] := -4; a2[4] := -5;
  var test2 := maxArray(a2);
  print("Test 2: maxArray([-1,-2,-3,-4,-5]) = ", test2, "\n");

  var a3 := new int[3];
  a3[0] := 0; a3[1] := 0; a3[2] := 0;
  var test3 := maxArray(a3);
  print("Test 3: maxArray([0,0,0]) = ", test3, "\n");

  var a4 := new int[2];
  a4[0] := 5; a4[1] := 10;
  var test4 := maxArray(a4);
  print("Test 4: maxArray([5,10]) = ", test4, "\n");

  var a5 := new int[1];
  a5[0] := 99;
  var test5 := maxArray(a5);
  print("Test 5: maxArray([99]) = ", test5, "\n");
}

method Main(){
  TestMethod();
}

\end{lstlisting}
\begin{tcolorbox}[title=\gpt Prompt,  label=systemprompt]
\small
\textbf{code2anno}: \\
\small{\texttt{You are an expert in Dafny. Fill in the weakest precondition and strongest postconditions for the dafny programs so that the dafny programs can be verified. \
    Do not change provided code. Exclude "requires true", "requires array!=null", "requires natural number >=0".\
    Do not assume input array or seq is non-empty. \
    Do not assume input integers are non-negative unless necessary. \
    Replace the //TOFILL string with the actual pre- and postconditions. Return the whole verifiable program.}}\\
    
\textbf{anno2code}: \\
\small{\texttt{You are an expert in dafny. You are given a dafny program with annotations.\
      Replace //TOFILL with the actual dafny code so that it can be verified. Return the whole program.\
      If loop is needed, use while instead of for. \
          Do not use helper functions.\
    DO NOT modify the function signature and annotations.}} \\

\textbf{doc2anno}: \\
\small{\texttt{You are an expert in Dafny. Generate the weakest preconditions and strongest postconditions for the dafny programs based on the docstring. \
    Do not change the provided code. Exclude "requires true", "requires array!=null", "requires natural number >=0".\
    Do not assume the input array or seq is non-empty. \
    Do not use self-defined functions. \
    Do not use int.MaxValue or int.MinValue. \
    Do not assume input integers are non-negative unless necessary.\
    Return only the annotations in code format starting with ``` and end with ```. Do not return method implementation.}}\\
    
\textbf{anno2doc}: \\
\small{\texttt{You are an expert in dafny. \
        Give one docstring of the given dafny annotation. Make sure to capture all details described in the annotation.}}\\
        
\textbf{code2doc}: \\
\small{\texttt{You are an expert in dafny. \
        Give one docstring of the given dafny code's functional behavior. Do not mention implementation details. Assume 'assert' as preconditions.}}\\
        
\textbf{doc2code}: \\
\small{\texttt{You are an expert in dafny. You are given a dafny program docstring.\
      Replace //TOFILL with the actual dafny code without annotation. Return the whole program.}}\\
      
\textbf{docstring equivalence checker}: \\
\small{\texttt{Determine if two docstrings describe the exact same functional behavior of a dafny program. \
        Make sure all details are exactly the same.}}

\end{tcolorbox}

\subsection{Wrong Ground-Truth programs in MBPP-DFY-50~\cite{misu2024towards}}
\label{sec:sup:mbpp_wrong}

Task 472 docstring states arrays can be empty or non-empty but the annotations require the array to be non-empty.
\begin{lstlisting}[style=dafnystyle, caption=mbpp-50 task 472,label=task472]

/* task_description: Write a method in Dafny to check whether the given integer array contains consecutive numbers or not.
preconditions: 
- There are no preconditions, the method will always work. Arrays can be empty or non-empty.
postconditions: 
- If the array contains consecutive numbers, the result is true
- If the array does not contain consecutive numbers, the result is false.
*/
method ContainsConsecutiveNumbers(a: array<int>) returns (result: bool)
    requires a.Length>0
    ensures result <==> (exists i :: 0 <= i < a.Length - 1 && a[i] + 1 == a[i + 1])
{
    result := false;
    for i := 0 to a.Length - 1
        invariant 0 <= i <= a.Length - 1
        invariant result <==> (exists k :: 0 <= k < i && a[k] + 1 == a[k + 1])
    {
        if a[i] + 1 == a[i + 1] {
            result := true;
            break;
        }
    }
}
\end{lstlisting}

Task 567 docstring states arrays can be empty or have any length but the annotations require the array to be non-empty.
\begin{lstlisting}[style=dafnystyle, caption=mbpp-50 task 567,label=task567]
/* task_description: Write a method in Dafny to check whether a specified array is sorted.
preconditions: 
- There are no preconditions, the method will always work. Arrays can be empty or have any length.
postconditions: 
- If the method returns true, the array is sorted in non-decreasing order.
- If the method returns false, the array is not sorted in non-decreasing order.
*/

method IsSorted(a: array<int>) returns (sorted: bool)
    requires a.Length > 0
    ensures sorted <== forall i, j :: 0 <= i < j < a.Length ==> a[i] <= a[j]
    ensures !sorted ==> exists i, j :: 0 <= i < j < a.Length && a[i] > a[j]
{
    sorted := true;
    for i := 0 to a.Length - 1
        invariant 0 <= i < a.Length
        invariant sorted <== forall k, l :: 0 <= k < l < i ==> a[k] <= a[l]
        invariant !sorted ==> exists k :: 0 <= k < i && a[k] > a[k+1]
    {
        if a[i] > a[i + 1]
        {
            sorted := false;
            break;
        }
    }
    sorted := sorted;
}
\end{lstlisting}

Task 576 annotations do not state the condition when the return is false.
\begin{lstlisting}[style=dafnystyle, caption=mbpp-50 task 576,label=task576]
/* task_description: Write a method in Dafny to check whether a list is sublist of another or not.
preconditions: 
- There are no preconditions, the method will always work. Sequences are always not null.
postconditions: 
- If the result is true, then the subsequence exists in the main sequence.
- If the result is false, then the subsequence does not exist in the main sequence.
*/
method IsSublist(sub: seq<int>, main: seq<int>) returns (result: bool)
    ensures true <== (exists i :: 0 <= i <= |main| - |sub| && sub == main[i..i + |sub|])
{
    if |sub| > |main| {
        return false;
    }

    for i := 0 to |main| - |sub| + 1
        invariant 0 <= i <= |main| - |sub| + 1
        invariant true <== (exists j :: 0 <= j < i && sub == main[j..j + |sub|])
    {
        if sub == main[i..i + |sub|] {
            result := true;
        }
    }
    result := false;
}
\end{lstlisting}

Task 632 docstring says there are no preconditions, but there is a \dafny annotation requiring the array to have length at least 2.  (Note that we rewrote the \verb|MoveZeroesToEnd| to get rid of the \verb|swap| helper method to run Clover consistency check, as \sys does not yet support helper methods.)
\begin{lstlisting}[style=dafnystyle, caption=mbpp-50 task 632,label=task632]
/* task_description: Write a method in Dafny to move all zeroes to the end of the given array.
preconditions: 
- There are no preconditions, the method will always work.
postconditions: 
- The length of the output array must be the same as the length of the input array.
- All zeroes in the input array are at the end of the output array.
- The relative order of the non-zero elements should be the same as in the input array.
- The number of zeroes in the input and output arrays should be the same.
*/

method MoveZeroesToEnd(arr: array<int>)
    requires arr.Length >= 2
    modifies arr
    // Same size
    ensures arr.Length == old(arr.Length)
    // Zeros to the right of the first zero
    ensures forall i, j :: 0 <= i < j < arr.Length && arr[i] == 0 ==> arr[j] == 0
    // The final array is a permutation of the original one
    ensures multiset(arr[..]) == multiset(old(arr[..]))
    // Relative order of non-zero elements is preserved
    ensures forall n, m /* on old array */:: 0 <= n < m < arr.Length && old(arr[n]) != 0 && old(arr[m]) != 0 ==> 
            exists k, l /* on new array */:: 0 <= k < l < arr.Length && arr[k] == old(arr[n]) && arr[l] == old(arr[m])
    //ensures IsOrderPreserved(arr[..], old(arr[..]))
    // Number of zeros is preserved
{
    var i := 0;
    var j := 0;

    assert 0 <= i  <= arr.Length;
    assert forall k :: 0 <= k < arr.Length ==> arr[k] == old(arr[k]);
    //assert(forall n, m :: 0 <= n < m < arr.Length  ==> arr[n] == old(arr[n]) && arr[m] == old(arr[m]));
    while j < arr.Length
        invariant 0 <= i <= j <= arr.Length
        // Elements to the right of j are unchanged
        invariant forall k :: j <= k < arr.Length ==> old(arr[k]) == arr[k]
        // Everything to the left of i is non-zero
        invariant forall k :: 0 <= k < i ==> arr[k] != 0
        // Everything between i and j, but excluding j, is zero
        invariant forall k :: i <= k < j ==> arr[k] == 0
        // If there there are zeros, they are to the right of i
        invariant forall k :: 0 <= k < j && arr[k] == 0 ==> k >= i
        // No new numbers are added, up to j
        invariant forall k :: 0 <= k < j && arr[k] != old(arr[k]) ==> exists l :: 0 <= l < j && arr[k] == old(arr[l])
        // The new array up to j is always a permutation of the original one
        invariant multiset(arr[..]) == multiset(old(arr[..]))
        // Relative order of non-zero elements is always preserved
        //invariant IsOrderPreserved(arr[..], old(arr[..]))
        invariant forall n, m /* on old */:: 0 <= n < m < j && old(arr[n]) != 0 && old(arr[m]) != 0 ==> 
            exists k, l /* on new */:: 0 <= k < l < i && arr[k] == old(arr[n]) && arr[l] == old(arr[m])
    {

        if arr[j] != 0
        {
            if i != j
            {
                assert(arr[j] != 0);
                swap(arr, i, j);
                assert(forall k :: 0 <= k <= j ==> exists l :: 0 <= l <= j && arr[k] == old(arr[l]));
            }
            i := i + 1;
        }
        j := j + 1;
    }
    assert j == arr.Length;
}

method swap(arr: array<int>, i: int, j: int)
    requires arr.Length > 0
    requires 0 <= i < arr.Length && 0 <= j < arr.Length
    modifies arr
    ensures arr[i] == old(arr[j]) && arr[j] == old(arr[i])
    ensures forall k :: 0 <= k < arr.Length && k != i && k != j ==> arr[k] == old(arr[k])
    ensures multiset(arr[..]) == multiset(old(arr[..]))
{
        var tmp := arr[i];
        arr[i] := arr[j];
        arr[j] := tmp;
}
\end{lstlisting}

Task 644 docstring states input k should be between 0 and the length of the array but the annotations state k is greater than or equal to 2.
\begin{lstlisting}[style=dafnystyle, caption=mbpp-50 task 644,label=task644]
/* task_description: Write a method in Dafny to reverse an array up to a given k.
preconditions: 
- k should be between 0 and the length of the array.
postconditions: 
- The input array is modified.
- The values of the array up to k are reversed.
- The values of the array after k remain unchanged.
*/

method ReverseUptoK(s: array<int>, k: int)
    modifies s
    requires 2 <= k <= s.Length
    ensures forall i :: 0 <= i < k ==> s[i] == old(s[k - 1 - i])
    ensures forall i :: k <= i < s.Length ==> s[i] == old(s[i])
{
	var l := k - 1;
	var i := 0;
	while (i < l-i)
		invariant 0 <= i <= (l+1)/2;
		invariant forall p :: 0 <= p < i || l-i < p <= l ==> s[p] == old(s[l-p]);
		invariant forall p :: i <= p <= l-i ==> s[p] == old(s[p]);
        invariant forall p :: k <= p < s.Length ==> s[p] == old(s[p])
	{
		s[i], s[l-i] := s[l-i], s[i];
		i := i + 1;
	}

}
\end{lstlisting}

Task 803 docstring states that if the result is false, there is no integer i such that i * i == n, but the corresponding annotation adds unnecessary bounds making the postcondition a tautology.
\begin{lstlisting}[style=dafnystyle, caption=mbpp-50 task 803,label=task803]
/* task_description: Write a method in Dafny to check whether the given number is a perfect square or not.
preconditions: 
-  n should be non-negative.
postconditions: 
- If the result is true, there exists an integer i such that i * i == n.
- If the result is false, there is no integer i such that i * i == n.
*/

method IsPerfectSquare(n: int) returns (result: bool)
    requires n >= 0
    ensures result == true ==> (exists i: int :: 0 <= i <= n && i * i == n)
    ensures result == false ==> (forall a: int :: 0 < a*a < n ==> a*a != n)
{
    var i := 0;
    while (i * i < n)
        invariant 0 <= i <= n
        invariant forall k :: 0 <= k < i ==> k * k < n
    {
        i := i + 1;
    }
    return i * i == n;
}
\end{lstlisting}
\newpage

\subsection{More Detailed Experiment Results}
\label{sec:appendix:detailed}
\newcounter{rowcount}
\begin{table}[ht!]
    \centering
    \setcounter{rowcount}{0}
    \resizebox{0.65\columnwidth}{!}{
\begin{tabular}{|c|c|c|c|c|c|}
    \hline
    Row No. & test & correct & correct\&accept & incorrect & incorrect\&accept \\ \hline

\stepcounter{rowcount}\arabic{rowcount} &abs&  20 & 20 & 0 & 0 \\ \hline
\stepcounter{rowcount}\arabic{rowcount} &all\_digits&  1 & 1 & 19 & 0 \\ \hline
\stepcounter{rowcount}\arabic{rowcount} &array\_append&  11 & 11 & 9 & 0 \\ \hline
\stepcounter{rowcount}\arabic{rowcount} &array\_concat&  1 & 1 & 19 & 0 \\ \hline
\stepcounter{rowcount}\arabic{rowcount} &array\_copy&  0 & 0 & 20 & 0 \\ \hline
\stepcounter{rowcount}\arabic{rowcount} &array\_product&  12 & 8 & 8 & 0 \\ \hline
\stepcounter{rowcount}\arabic{rowcount} &array\_sum&  2 & 1 & 18 & 0 \\
\hline
\end{tabular}}
    \caption{End2End generation}
    \label{table:end2end}
\end{table}

\begin{table}
    \centering
    \setcounter{rowcount}{0}
    \resizebox{0.95\columnwidth}{!}{
    \begin{tabular}{|c|c|c|c|c|c|c|c|}
    \hline
    Row No. & test & anno-complete & doc2anno & doc2code & anno2doc & code2doc & 3-edges\\ \hline

\stepcounter{rowcount}\arabic{rowcount} &abs &  A &   A &   A &  A &  A &  A  \\ \hline
\stepcounter{rowcount}\arabic{rowcount} &array\_append &  A &   R &   A &  A &  A &  R  \\ \hline
\stepcounter{rowcount}\arabic{rowcount} &array\_concat &  A &   R &   A &  A &  A &  R  \\ \hline
\stepcounter{rowcount}\arabic{rowcount} &array\_copy &  A &   A &   A &  A &  A &  A  \\ \hline
\stepcounter{rowcount}\arabic{rowcount} &array\_product &  A &   R &   A &  A &  A &  R  \\ \hline
\stepcounter{rowcount}\arabic{rowcount} &array\_sum &  A &   A &   A &  A &  A &  A  \\ \hline
\stepcounter{rowcount}\arabic{rowcount} &avg &  A &   A &   A &  A &  A &  A  \\
\hline

    \end{tabular}}
    \caption{Verus CLOVER ground truth experiments with k=1}
    \label{table:combined_table_k}
\end{table}

\begin{table}
    \centering
    \setcounter{rowcount}{0}
    \resizebox{0.95\columnwidth}{!}{
    \begin{tabular}{|c|c|c|c|c|c|c|c|}
    \hline
    Row No. & test & anno-complete & doc2anno & doc2code & anno2doc & code2doc & 3-edges\\ \hline

\stepcounter{rowcount}\arabic{rowcount} &abs &  A &   A &   A &  A &  A &  A  \\ \hline
\stepcounter{rowcount}\arabic{rowcount} &array\_append &  A &   A &   A &  A &  A &  A  \\ \hline
\stepcounter{rowcount}\arabic{rowcount} &array\_concat &  A &   A &   A &  A &  A &  A  \\ \hline
\stepcounter{rowcount}\arabic{rowcount} &array\_copy &  A &   A &   A &  A &  A &  A  \\ \hline
\stepcounter{rowcount}\arabic{rowcount} &array\_product &  A &   A &   A &  A &  A &  A  \\ \hline
\stepcounter{rowcount}\arabic{rowcount} &array\_sum &  A &   A &   A &  A &  A &  A  \\ \hline
\stepcounter{rowcount}\arabic{rowcount} &avg &  A &   A &   A &  A &  A &  A  \\
\hline

    \end{tabular}}
    \caption{Verus CLOVER ground truth experiments with k=10}
    \label{table:combined_table_k}
\end{table}

\begin{table}
    \centering
    \setcounter{rowcount}{0}
    \resizebox{0.85\columnwidth}{!}{
    \begin{tabular}{|c|c|c|c|c|c|c|c|}
    \hline
    Row No. & test & anno-complete & doc2anno & doc2code & anno2doc & code2doc & 3-edges\\ \hline
\stepcounter{rowcount}\arabic{rowcount} &abs & A  & A  & A  & A  & A  & A  \\ \hline
\stepcounter{rowcount}\arabic{rowcount} &all\_digits & A  & A  & A  & A  & A  & A  \\ \hline
\stepcounter{rowcount}\arabic{rowcount} &array\_append & A  & A  & A  & A  & A  & A  \\ \hline
\stepcounter{rowcount}\arabic{rowcount} &array\_concat & A  & A  & A  & A  & A  & A  \\ \hline
\stepcounter{rowcount}\arabic{rowcount} &array\_copy & A  & A  & A  & A  & A  & A  \\ \hline
\stepcounter{rowcount}\arabic{rowcount} &array\_product & A  & A  & A  & A  & A  & A  \\ \hline
\stepcounter{rowcount}\arabic{rowcount} &array\_sum & A  & A  & A  & A  & A  & A  \\
\hline
    \end{tabular}}
    \caption{\small CLOVER ground truth experiments with k=10: Each row represents one example. Each column represents one directed edge. In each cell, there is either \textbf{A} or \textbf{R}.  \textbf{A} means that the directed edge is accepted by our checker and  \textbf{R} means that the cell is rejected by our checker. In each row, if all edges are accepted, then the example is accepted.}
    \label{table:combined_table_k}
\end{table}

\begin{table}
    \centering
    \setcounter{rowcount}{0}
    \resizebox{0.95\columnwidth}{!}{
    \begin{tabular}{|c|c|c|c|c|c|c|c|}
    \hline
    Row No. & test & anno-complete & doc2anno & doc2code & anno2doc & code2doc & 3-edges\\ \hline
\stepcounter{rowcount}\arabic{rowcount} &abs & A  & A  & A  & A  & A  & A  \\ \hline
\stepcounter{rowcount}\arabic{rowcount} &all\_digits & A  & A  & A  & A  & A  & A  \\ \hline
\stepcounter{rowcount}\arabic{rowcount} &array\_append & A  & A  & A  & A  & A  & A  \\ \hline
\stepcounter{rowcount}\arabic{rowcount} &array\_concat & A  & A  & A  & A  & A  & A  \\ \hline
\stepcounter{rowcount}\arabic{rowcount} &array\_copy & A  & A  & A  & A  & A  & A  \\ \hline
\stepcounter{rowcount}\arabic{rowcount} &array\_product & A  & A  & A  & A  & A  & A  \\ \hline
\stepcounter{rowcount}\arabic{rowcount} &array\_sum & A  & A  & A  & A  & A  & A  \\
\hline
    \end{tabular}}
    \caption{CLOVER ground truth experiments when k=1}
    \label{table:merged}
\end{table}

\begin{table}

\footnotesize
    \centering
    \setcounter{rowcount}{0}
    \resizebox{0.95\columnwidth}{!}{
    \begin{tabular}{|c|c|c|c|c|c|c|}
    \hline
    Row No. & test & max3tries & oneTry & noVerify\_max3tries & withDoc\_max3tries & max3tries\_Claude \\ \hline
\stepcounter{rowcount}\arabic{rowcount} &abs & A  & A  & A  & A  & A  \\ \hline
\stepcounter{rowcount}\arabic{rowcount} &all\_digits & A  & A  & A  & A  & R  \\ \hline
\stepcounter{rowcount}\arabic{rowcount} &array\_append & A  & A  & A  & A  & A  \\ \hline
\stepcounter{rowcount}\arabic{rowcount} &array\_concat & A  & A  & A  & A  & A  \\ \hline
\stepcounter{rowcount}\arabic{rowcount} &array\_copy & A  & A  & A  & A  & A  \\ \hline
\stepcounter{rowcount}\arabic{rowcount} &array\_product & A  & R  & A  & A  & A  \\ \hline
\stepcounter{rowcount}\arabic{rowcount} &array\_sum & A  & R  & A  & A  & A  \\
\hline
\end{tabular}}
    \caption{\small Ablation studies that compare code generation under different configurations. Each column represents one configuration. We have: max3tries (a maximum of 3 tries with verifier feedback), oneTry (the first try), noVerify\_max3tries (a maximum of 3 tries with only compiler and no verifier feedback), withDoc\_max3tries (a maximum of 3 tries with verifier feedback plus docstrings), and max3tries\_Claude (same as max3tries using Claude API).}
    \label{table:anno2code_combined_table}
\end{table}

\begin{table}
    \centering
\setcounter{rowcount}{0}
    \resizebox{0.45\textwidth}{!}{
\begin{tabular}{|c|c|c|c|}
    \hline
    Row No. & test & max3tries & oneTry \\ \hline
\stepcounter{rowcount}\arabic{rowcount} &abs & A  & A  \\ \hline
\stepcounter{rowcount}\arabic{rowcount} &all\_digits & A  & A  \\ \hline
\stepcounter{rowcount}\arabic{rowcount} &array\_append & A  & A  \\ \hline
\stepcounter{rowcount}\arabic{rowcount} &array\_concat & A  & A \\ \hline
\stepcounter{rowcount}\arabic{rowcount} &array\_copy & A  & A  \\ \hline
\stepcounter{rowcount}\arabic{rowcount} &array\_product & A  & A  \\ \hline
\stepcounter{rowcount}\arabic{rowcount} &array\_sum & A  & R  \\
\hline
\end{tabular}}
    \caption{\small Ablation studies for annotation generation from pure code: note that this is the only place where we count loop invariants in annotations.  \textbf{A} means (1) generated annotations are equivalent to the original and (2) generated loop invariants are enough to have the code verified by \dafny. }
    \label{table:code2anno_feedback_ab}
\end{table}

\begin{table}
    \centering
    \setcounter{rowcount}{0}
    \resizebox{1\columnwidth}{!}{
    \begin{tabular}{|c|c|c|c|c|c|c|c|}
    \hline
    Row No. & test & anno-sound & anno-complete & doc2anno & doc2code & anno2doc & code2doc\\ \hline

\stepcounter{rowcount}\arabic{rowcount} &abs & A  & A  & R  & R  & R  & R  \\ \hline
\stepcounter{rowcount}\arabic{rowcount} &all\_digits & A  & A  & R  & R  & R  & R  \\ \hline
\stepcounter{rowcount}\arabic{rowcount} &array\_append & A  & A  & R  & A  & R  & R  \\ \hline
\stepcounter{rowcount}\arabic{rowcount} &array\_concat & A  & A  & R  & A  & A  & R  \\ \hline
\stepcounter{rowcount}\arabic{rowcount} &array\_copy & A  & A  & R  & A  & R  & R  \\ \hline
\stepcounter{rowcount}\arabic{rowcount} &array\_product & A  & A  & R  & A  & R  & R  \\ \hline
\stepcounter{rowcount}\arabic{rowcount} &array\_sum & A  & A  & A  & A  & A  & R  \\
\hline

\end{tabular}}
    \caption{C1 when k=1: all examples are rejected}
    \label{table:new_C1_k1}
\end{table}

\begin{table}
    \centering
    \setcounter{rowcount}{0}
    \resizebox{1\columnwidth}{!}{
    \begin{tabular}{|c|c|c|c|c|c|c|c|}
    \hline
    Row No. & test & anno-sound & anno-complete & doc2anno & doc2code & anno2doc & code2doc\\ \hline

\stepcounter{rowcount}\arabic{rowcount} &abs & A  & A  & R  & R  & R  & R  \\ \hline
\stepcounter{rowcount}\arabic{rowcount} &all\_digits & A  & A  & R  & R  & R  & R  \\ \hline
\stepcounter{rowcount}\arabic{rowcount} &array\_append & A  & A  & R  & A  & R  & R  \\ \hline
\stepcounter{rowcount}\arabic{rowcount} &array\_concat & A  & A  & R  & A  & A  & R  \\ \hline
\stepcounter{rowcount}\arabic{rowcount} &array\_copy & A  & A  & R  & A  & R  & R  \\ \hline
\stepcounter{rowcount}\arabic{rowcount} &array\_product & A  & A  & R  & A  & A  & R  \\ \hline
\stepcounter{rowcount}\arabic{rowcount} &array\_sum & A  & A  & A  & A  & A  & R  \\
\hline
\end{tabular}}
    \caption{C1 when k=10: all rejected}
    \label{table:new_C1_k10}
\end{table}

\begin{table}
    \centering
    \setcounter{rowcount}{0}
    \resizebox{1\columnwidth}{!}{
    \begin{tabular}{|c|c|c|c|c|c|c|c|}
    \hline
    Row No. & test & anno-sound & anno-complete & doc2anno & doc2code & anno2doc & code2doc\\ \hline
\stepcounter{rowcount}\arabic{rowcount} &abs & A  & R  & R  & A  & R  & A  \\ \hline
\stepcounter{rowcount}\arabic{rowcount} &all\_digits & A  & R  & R  & A  & A  & A  \\ \hline
\stepcounter{rowcount}\arabic{rowcount} &array\_append & A  & A  & R  & A  & R  & A  \\ \hline
\stepcounter{rowcount}\arabic{rowcount} &array\_concat & A  & A  & R  & A  & A  & A  \\ \hline
\stepcounter{rowcount}\arabic{rowcount} &array\_copy & A  & A  & R  & A  & R  & A  \\ \hline
\stepcounter{rowcount}\arabic{rowcount} &array\_product & A  & A  & R  & A  & A  & A  \\ \hline
\stepcounter{rowcount}\arabic{rowcount} &array\_sum & A  & A  & R  & A  & R  & A  \\
\hline
\end{tabular}}
    \caption{C2 when k=1: all examples are rejected}
    \label{table:new_C2_k1}
\end{table}

\begin{table}
    \centering
    \setcounter{rowcount}{0}
    \resizebox{1\columnwidth}{!}{
    \begin{tabular}{|c|c|c|c|c|c|c|c|}
    \hline
    Row No. & test & anno-sound & anno-complete & doc2anno & doc2code & anno2doc & code2doc\\ \hline
\stepcounter{rowcount}\arabic{rowcount} &abs & A  & A  & R  & A  & R  & A  \\ \hline
\stepcounter{rowcount}\arabic{rowcount} &all\_digits & A  & R  & R  & A  & A  & A  \\ \hline
\stepcounter{rowcount}\arabic{rowcount} &array\_append & A  & A  & R  & A  & R  & A  \\ \hline
\stepcounter{rowcount}\arabic{rowcount} &array\_concat & A  & A  & R  & A  & A  & A  \\ \hline
\stepcounter{rowcount}\arabic{rowcount} &array\_copy & A  & A  & R  & A  & R  & A  \\ \hline
\stepcounter{rowcount}\arabic{rowcount} &array\_product & A  & A  & R  & A  & A  & A  \\ \hline
\stepcounter{rowcount}\arabic{rowcount} &array\_sum & A  & A  & R  & A  & A  & A  \\
\hline
\end{tabular}}
    \caption{C2 when k=10: all examples are rejected}
    \label{table:new_C2_k10}
\end{table}

\begin{table}
    \centering
    \setcounter{rowcount}{0}
    \resizebox{1\columnwidth}{!}{
    \begin{tabular}{|c|c|c|c|c|c|c|c|}
    \hline
    Row No. & test & anno-sound & anno-complete & doc2anno & doc2code & anno2doc & code2doc\\ \hline

\stepcounter{rowcount}\arabic{rowcount} &abs & A  & R  & A  & R  & R  & R  \\ \hline
\stepcounter{rowcount}\arabic{rowcount} &all\_digits & A  & R  & R  & R  & A  & A  \\ \hline
\stepcounter{rowcount}\arabic{rowcount} &array\_append & A  & A  & R  & A  & R  & R  \\ \hline
\stepcounter{rowcount}\arabic{rowcount} &array\_concat & A  & A  & R  & A  & R  & R  \\ \hline
\stepcounter{rowcount}\arabic{rowcount} &array\_copy & A  & A  & A  & A  & R  & R  \\ \hline
\stepcounter{rowcount}\arabic{rowcount} &array\_product & A  & A  & R  & R  & A  & R  \\ \hline
\stepcounter{rowcount}\arabic{rowcount} &array\_sum & A  & A  & R  & A  & R  & R  \\
\hline
\end{tabular}}
    \caption{C3 when k=1: all examples are rejected}
    \label{table:new_C3_k1}
\end{table}

\begin{table}
    \centering
    \setcounter{rowcount}{0}
    \resizebox{1\columnwidth}{!}{
    \begin{tabular}{|c|c|c|c|c|c|c|c|}
    \hline
    Row No. & test & anno-sound & anno-complete & doc2anno & doc2code & anno2doc & code2doc\\ \hline

\stepcounter{rowcount}\arabic{rowcount} &abs & A  & A  & A  & R  & A  & R  \\ \hline
\stepcounter{rowcount}\arabic{rowcount} &all\_digits & A  & R  & R  & R  & A  & A  \\ \hline
\stepcounter{rowcount}\arabic{rowcount} &array\_append & A  & A  & R  & A  & A  & R  \\ \hline
\stepcounter{rowcount}\arabic{rowcount} &array\_concat & A  & A  & R  & A  & R  & A  \\ \hline
\stepcounter{rowcount}\arabic{rowcount} &array\_copy & A  & A  & A  & A  & A  & R  \\ \hline
\stepcounter{rowcount}\arabic{rowcount} &array\_product & A  & A  & R  & A  & A  & R  \\ \hline
\stepcounter{rowcount}\arabic{rowcount} &array\_sum & A  & A  & A  & A  & A  & R  \\
\hline

\end{tabular}}
    \caption{C3 when k=10: all examples are rejected}
    \label{table:new_C3_k10}
\end{table}

\begin{table}
    \centering
    \setcounter{rowcount}{0}
    \resizebox{1\columnwidth}{!}{
    \begin{tabular}{|c|c|c|c|c|c|c|c|}
    \hline
    Row No. & test & anno-sound & anno-complete & doc2anno & doc2code & anno2doc & code2doc\\ \hline

\stepcounter{rowcount}\arabic{rowcount} &abs & A  & A  & R  & R  & R  & R  \\ \hline
\stepcounter{rowcount}\arabic{rowcount} &all\_digits & A  & R  & R  & R  & A  & A  \\ \hline
\stepcounter{rowcount}\arabic{rowcount} &array\_append & A  & R  & R  & R  & A  & A  \\ \hline
\stepcounter{rowcount}\arabic{rowcount} &array\_concat & A  & A  & R  & R  & R  & R  \\ \hline
\stepcounter{rowcount}\arabic{rowcount} &array\_copy & A  & R  & R  & R  & R  & R  \\ \hline
\stepcounter{rowcount}\arabic{rowcount} &array\_product & A  & A  & R  & R  & R  & R  \\ \hline
\stepcounter{rowcount}\arabic{rowcount} &array\_sum & A  & R  & R  & R  & R  & R  \\
\hline
\end{tabular}}
    \caption{C6 when k=1: all examples are rejected}
    \label{table:new_C6_k1}
\end{table}

\begin{table}
    \centering
    \setcounter{rowcount}{0}
    \resizebox{1\columnwidth}{!}{
    \begin{tabular}{|c|c|c|c|c|c|c|c|}
    \hline
    Row No. & test & anno-sound & anno-complete & doc2anno & doc2code & anno2doc & code2doc\\ \hline

\stepcounter{rowcount}\arabic{rowcount} &abs & A  & A  & R  & R  & R  & R  \\ \hline
\stepcounter{rowcount}\arabic{rowcount} &all\_digits & A  & R  & R  & R  & A  & A  \\ \hline
\stepcounter{rowcount}\arabic{rowcount} &array\_append & A  & R  & R  & R  & A  & A  \\ \hline
\stepcounter{rowcount}\arabic{rowcount} &array\_concat & A  & A  & R  & R  & R  & R  \\ \hline
\stepcounter{rowcount}\arabic{rowcount} &array\_copy & A  & R  & R  & R  & R  & R  \\ \hline
\stepcounter{rowcount}\arabic{rowcount} &array\_product & A  & A  & R  & R  & R  & R  \\ \hline
\stepcounter{rowcount}\arabic{rowcount} &array\_sum & A  & R  & R  & R  & A  & R  \\
\hline
\end{tabular}}
    \caption{C6 when k=10: all examples are rejected }
    \label{table:new_C6_k10}
\end{table}

\begin{table}
    \centering
    \setcounter{rowcount}{0}
    \resizebox{0.85\columnwidth}{!}{
    \begin{tabular}{|c|c|c|c|c|c|c|c|}
    \hline
    Row No. & test & doc2code & code2doc & doc2anno & anno2doc & anno-sound & anno-complete\\ \hline
\stepcounter{rowcount}\arabic{rowcount} &task\_id\_101 & A  & A  & A  & A  & A  & A  \\ \hline
\stepcounter{rowcount}\arabic{rowcount} &task\_id\_267 & A  & A  & A  & A  & A  & A  \\ \hline
\stepcounter{rowcount}\arabic{rowcount} &task\_id\_404 & A  & A  & A  & A  & A  & A  \\ \hline
\stepcounter{rowcount}\arabic{rowcount} &task\_id\_770 & A  & A  & R  & A  & A  & A  \\ \hline
\stepcounter{rowcount}\arabic{rowcount} &task\_id\_599 & A  & A  & A  & A  & A  & A  \\ \hline
\stepcounter{rowcount}\arabic{rowcount} &task\_id\_610 & A  & A  & A  & A  & A  & A  \\ \hline
\stepcounter{rowcount}\arabic{rowcount} &task\_id\_605 & A  & A  & A  & A  & A  & A  \\
\hline
    \end{tabular}}
    \caption{MBPP-DFY-50 ground truth results when k=10}
    \label{table:mbpp_table}
\end{table}

\end{document}